\documentclass[10pt,twocolumn,letterpaper,table]{article}

\usepackage[pagenumbers]{cvpr} 
\usepackage{multirow}
\usepackage{xcolor}

\definecolor{cvprblue}{rgb}{0.21,0.49,0.74}
\usepackage[pagebackref,breaklinks,colorlinks,allcolors=cvprblue]{hyperref}

\newcommand{\name}{STAR}
\title{STAR: Spatial-Temporal Augmentation with Text-to-Video Models\\ for Real-World Video Super-Resolution}

\author{
  Rui Xie$^{1*}$, \hspace{0.2cm}
  Yinhong Liu$^{1*}$, \hspace{0.2cm}
  Penghao Zhou$^2$, \hspace{0.2cm}
  Chen Zhao$^1$, \hspace{0.2cm}
  Jun Zhou$^3$ \\
  Kai Zhang$^1$, \hspace{0.2cm}
  Zhenyu Zhang$^1$, \hspace{0.2cm}
  Jian Yang$^{1}$, \hspace{0.2cm}
  Zhenheng Yang$^2$, \hspace{0.2cm}
  Ying Tai$^{1\dagger}$ \\
  $^1$Nanjing University, \hspace{0.2cm}
  $^2$ByteDance, \hspace{0.2cm}
  $^3$Southwest University \\
  {\small \url{https://nju-pcalab.github.io/projects/STAR}}
}

\begin{document}

\twocolumn[{
\renewcommand\twocolumn[1][]{#1}
\maketitle
\begin{center}
    \captionsetup{type=figure}
    \includegraphics[width=\textwidth]{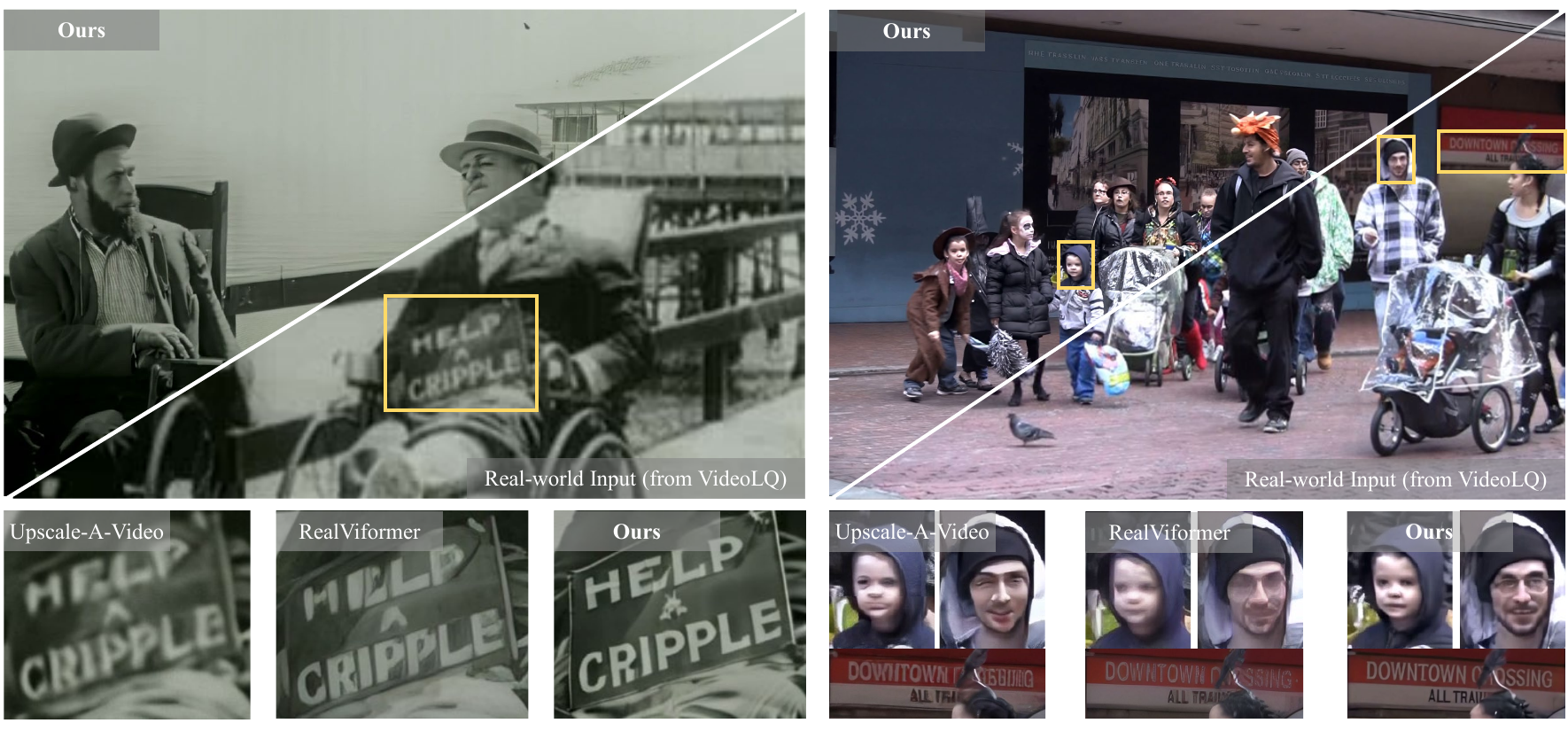} \vspace{-6mm}
    \captionof{figure}{Visualization comparisons on both real-world and synthetic low-resolution videos. 
    Compared to the state-of-the-art VSR models~\cite{zhang2024realviformer,zhou2024upscale}, our results demonstrate more natural facial details and better structure of the text. 
    \textbf{(Zoom-in for best view)}}
    \label{teaser}
\end{center}
}]

\begin{abstract}
Image diffusion models have been adapted for real-world video super-resolution to tackle over-smoothing issues in GAN-based methods.
However, these models struggle to maintain temporal consistency, as they are trained on static images, limiting their ability to capture temporal dynamics effectively.
Integrating text-to-video (T2V) models into video super-resolution for improved temporal modeling is straightforward. 
However, two key challenges remain: artifacts introduced by complex degradations in real-world scenarios, and compromised fidelity due to the strong generative capacity of powerful T2V models (\textit{e.g.}, CogVideoX-5B). 
To enhance the spatio-temporal quality of restored videos, we introduce\textbf{~\name} (\textbf{S}patial-\textbf{T}emporal \textbf{A}ugmentation with T2V models for \textbf{R}eal-world video super-resolution), a novel approach that leverages T2V models for real-world video super-resolution, achieving realistic spatial details and robust temporal consistency.
Specifically, we introduce a Local Information Enhancement Module (LIEM) before the global attention block to enrich local details and mitigate degradation artifacts. 
Moreover, we propose a Dynamic Frequency (DF) Loss to reinforce fidelity, guiding the model to focus on different frequency components across diffusion steps. 
Extensive experiments demonstrate\textbf{~\name}~outperforms state-of-the-art methods on both synthetic and real-world datasets.\footnotemark
\footnotetext{$^*$Equal contributions. Work done during Rui Xie’s ByteDance internship. $\dagger$ indicates corresponding author.}
\end{abstract}
\vspace{-4mm}
\section{Introduction}
\label{sec:intro}
Real-world video super-resolution (VSR) aims to generate high-resolution (HR) videos with clear details and strong temporal consistency from low-resolution (LR) inputs with unknown degradations. 
Most VSR methods \cite{wang2019edvr, chan2022basicvsr++, jo2018deep, xue2019video} only focus on simple, known degradations like downsampling \cite{fuoli2019efficient, isobe2020video} or camera-related issues~\cite{yang2021realvsr}. 
However, real-world scenarios often involve \textit{unexpected degradations} such as noise, blur, and compression, making it difficult for models to capture both spatial and temporal information needed for high-quality, consistent restoration.

GAN-based methods \cite{zhang2024realviformer, chan2022investigating, wang2021realesrgan, wu2022animesr, yang2021realvsr} are widely used in real-world VSR for improving details through adversarial learning. 
By incorporating optical flow maps, they also improve temporal consistency, yielding smooth motion across frames. However, their limited generative capacity often results in oversmoothing, as illustrated in Figure~\ref{teaser}.
Recently, image diffusion models \cite{rombach2022high} have been applied to real-world VSR for realistic video generation. Methods like \cite{zhou2024upscale, chen2024learning, yuan2024inflation, yang2023mgldvsr} incorporate temporal blocks or optical flow maps to improve temporal information capture. However, since these models are primarily trained on image data rather than video data~\cite{nan2024openvid, chen2024panda, wang2023internvid, wang2024vidprom}, simply adding temporal layers often fails to ensure high temporal consistency (see Figure~\ref{fig:temp_consis}). 
VEnhancer \cite{he2024venhancer} and LaVie-SR \cite{wang2023lavie} incorporate T2V models for super-resolving AI-generated videos. However, two key challenges still remain: \textit{artifacts introduced by complex degradations} in real-world settings, and \textit{compromised fidelity} due to the strong generative capacity of powerful T2V models (\textit{e.g.}, CogVideoX).

To fully leverage the T2V prior~\cite{zhang2023i2vgen, yang2024cogvideox} to enhance practical VSR, we introduce~\name, a novel Spatial-Temporal Augmentation approach for Real-world VSR that achieves realistic spatial details and robust temporal consistency.
Specifically,
$1$) To address artifacts, we introduce a Local Information Enhancement Module (LIEM) before global self-attention to evaluate its impact on T2V models for real-world VSR. 
This approach stems from our observation that most T2V models rely solely on a global information extraction module (\textit{i.e.}, global self-attention), whereas capturing local details is crucial for video restoration.
$2$) To improve fidelity, we propose a Dynamic Frequency (DF) Loss, guiding the model to prioritize low- or high-frequency information at different diffusion steps. 
This is based on our observation that during the reverse diffusion process, our model tends to first recover structure and then refine details. This approach decouples fidelity requirements, reduces learning difficulty, and enhances restoration fidelity.

In summary, our main contributions are as follows:

$\bullet$ 
We propose~\name, a Spatio-Temporal quality Augmentation framework for Real-world VSR.
To our best knowledge, we are the first to integrate diverse, powerful text-to-video diffusion priors into real-world VSR, improving both spatial details and temporal consistency.

$\bullet$ 
We introduce LIEM to enhance local details and ease degradation removal, effectively mitigating artifacts. 
Moreover, we propose DF loss to guide the model in learning frequency-specific information across diffusion steps, decoupling fidelity requirements and ultimately improving overall fidelity.

$\bullet$ Our~\name~achieves the highest clarity (DOVER scores) across all datasets compared to state-of-the-art methods, while maintaining robust temporal consistency.
\section{Related Work}
\label{sec:related work}

\begin{figure*}[t]
    \centering
    \includegraphics[width=1\linewidth]{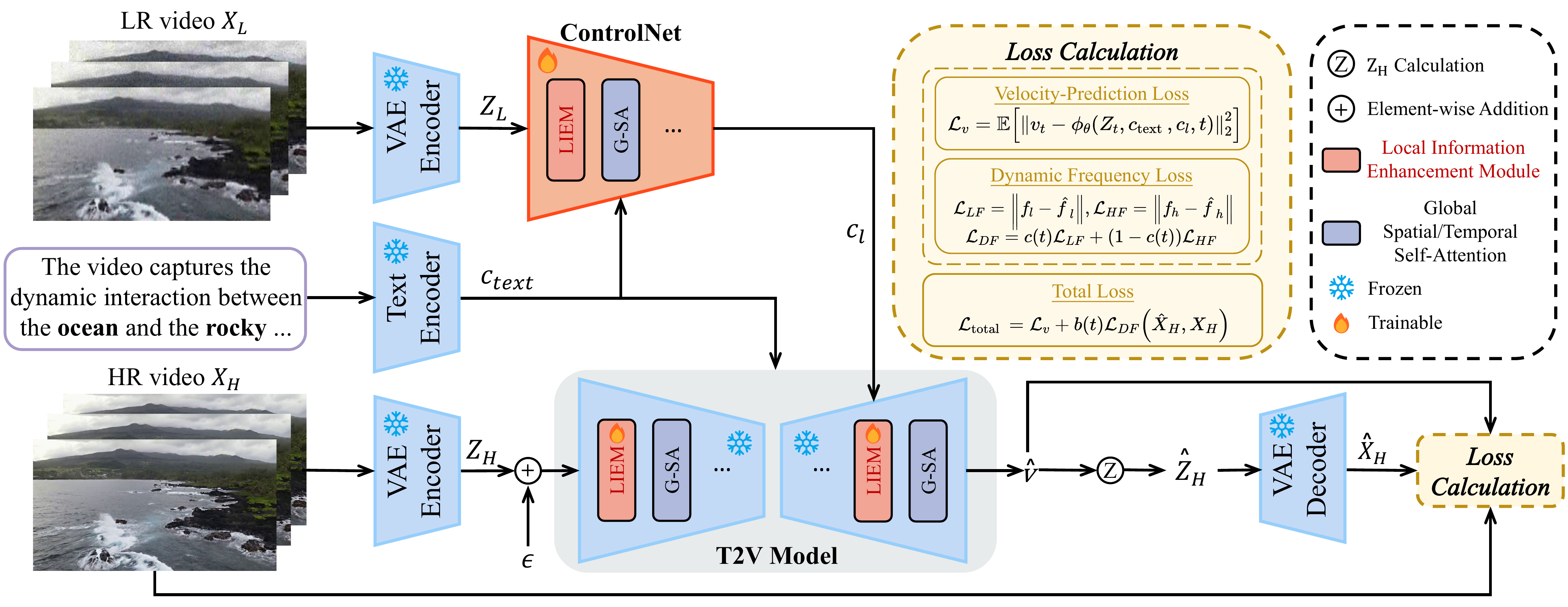}\vspace{-2mm}
    \caption{Overview of the proposed~\name.}
    \label{fig:overview}
\end{figure*}

\paragraph{Video Super-Resolution.} 
Traditional VSR methods can be roughly divided into two categories: recurrent-based \cite{haris2019recurrent, huang2017video, liang2022recurrent, sajjadi2018frame, shi2022rethinking} and sliding-window-based \cite{caballero2017real,liang2024vrt,li2020mucan,xu2021temporal,yi2019progressive} methods. 
Recurrent-based methods process LR video frame by frame using recurrent neural networks \cite{mikolov2010recurrent}. In contrast, sliding-window-based methods divide a video sequence into segments, using each as input to super-resolve the video. 
However, both approaches suffer from degradation mismatch, leading to significant performance drops in real-world applications.
Recently, there has been a growing focus on real-world VSR, targeting complex, unknown degradations. RealBasicVSR \cite{chan2022investigating}, an extension of BasicVSR \cite{chan2021basicvsr}, introduces a pre-cleaning module to mitigate artifacts. 
RealViformer \cite{zhang2024realviformer} discovers that channel attention is less sensitive to artifacts and uses squeeze-excite mechanisms and covariance-based rescaling to address these challenges further. While GAN-based and image diffusion models have made substantial progress, they still face issues such as \textit{over-smoothing details and temporal inconsistency}.

\vspace{-1em}
\paragraph{Text-to-Video Diffusion Model.} 
Large-scale pre-trained text-to-video (T2V) diffusion models have garnered significant attention, particularly with the impressive results from Sora \cite{videoworldsimulators2024,sora}. 
Numerous T2V models have since emerged, generally divided into: U-Net-based methods \cite{blattmann2023align, blattmann2023stable, ho2022imagen, singer2022make} and DiT-based methods \cite{yang2024cogvideox, bao2024vidu, polyak2024movie, chen2024gentron}. 
I2VGen-XL \cite{zhang2023i2vgen}, a U-Net-based method, employs a two-stage approach: first generating semantically and content-consistent LR videos, then using these as conditions to produce HR outputs.
CogvideoX \cite{yang2024cogvideox}, built on DiT \cite{peebles2023scalable}, introduces an adaptive LayerNorm to enhance text-video alignment and employs 3D attention to better integrate spatio-temporal information.
Both models 
have large model capacities and are trained on large-scale datasets, enabling them to capture robust spatio-temporal priors. 
In this work, we propose~\name~to \textit{fully leverage T2V model prior for real-world VSR}.

\vspace{-1em}
\paragraph{Diffusion Prior for Super-Resolution.}
Several works \cite{wang2024exploiting, lin2023diffbir, yang2023pixel, wu2024seesr, zhao2024wavelet} have leveraged generative diffusion priors for image and video super-resolution. 
StableSR \cite{wang2024exploiting} adds a time-aware encoder and feature warping module to the SD model. DiffBIR \cite{lin2023diffbir} integrates restoration and generative modules via ControlNet, while PASD \cite{yang2023pixel} and SeeSR \cite{wu2024seesr} embed semantic information in U-Net to guide diffusion. These methods balance fidelity and perceptual quality, achieving high-resolution image details.
Methods like Upscale-A-Video \cite{zhou2024upscale}, MGLD-VSR \cite{yang2023mgldvsr}, Inflating with Diffusion \cite{yuan2024inflation}, and SATeCo \cite{chen2024learning} have adapted text-to-image diffusion priors~\cite{rombach2022high,ho2022imagen} for VSR by adding temporal layers. However, rooted in text-to-image models, they often struggle with temporal consistency. 
More recently, VEnhancer\cite{he2024venhancer} and LaVie-SR\cite{wang2023lavie} have incorporated T2V models to super-resolve AI-generated videos but struggle with complex degradations in practical environments. 
In contrast, we are the \textit{first} to integrate powerful T2V diffusion priors for real-world VSR, introducing the LIEM module to address spatial artifacts and DF loss to enhance fidelity.
\section{Methodology}
\label{sec:methodology}

\begin{figure*}[t!]
    \centering
    \includegraphics[width=0.98\linewidth]{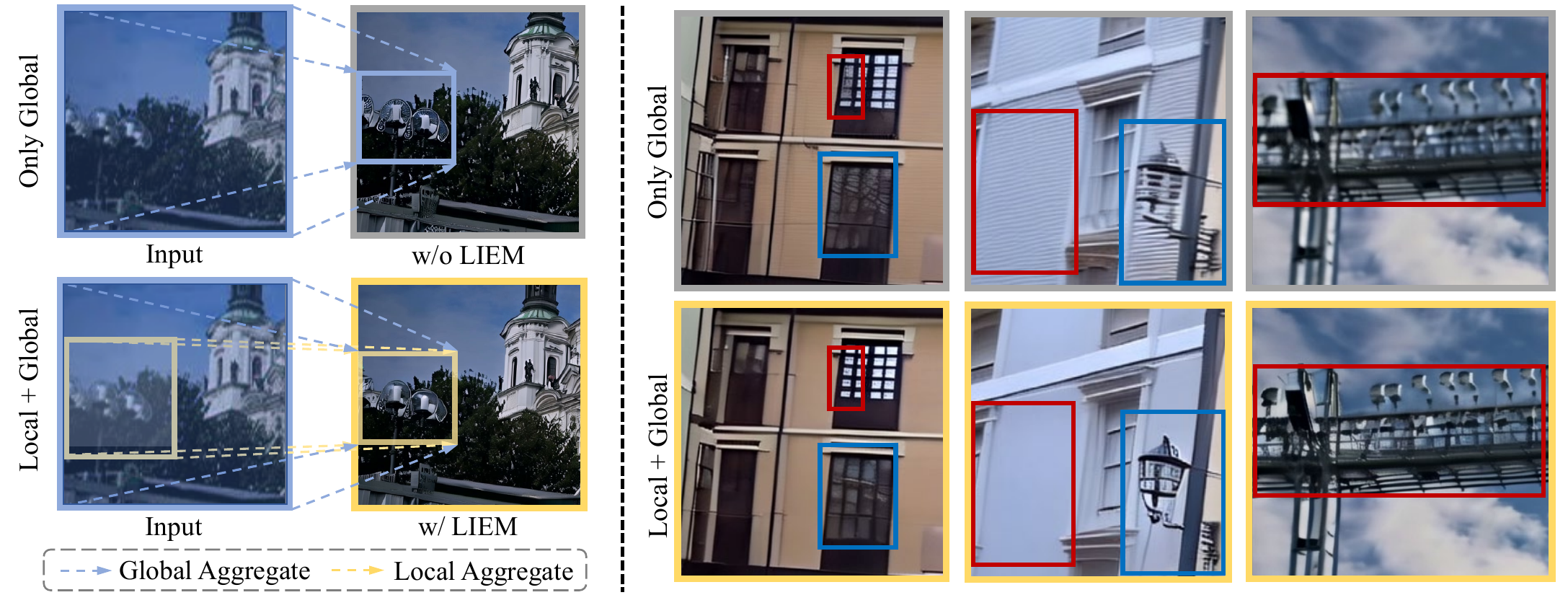}\vspace{-2mm}
    \caption{\textbf{Motivation of LIEM.} \textbf{Left:} schematic diagram illustrating the impact of using only global structure versus a combination of local and global structures. \textbf{Right:} visual comparison on real-world and synthetic videos. (\textbf{Zoom-in for best view})}
    \label{motivation:liem}
\end{figure*}

\begin{figure*}[t!]
    \centering
    \includegraphics[width=1\linewidth]{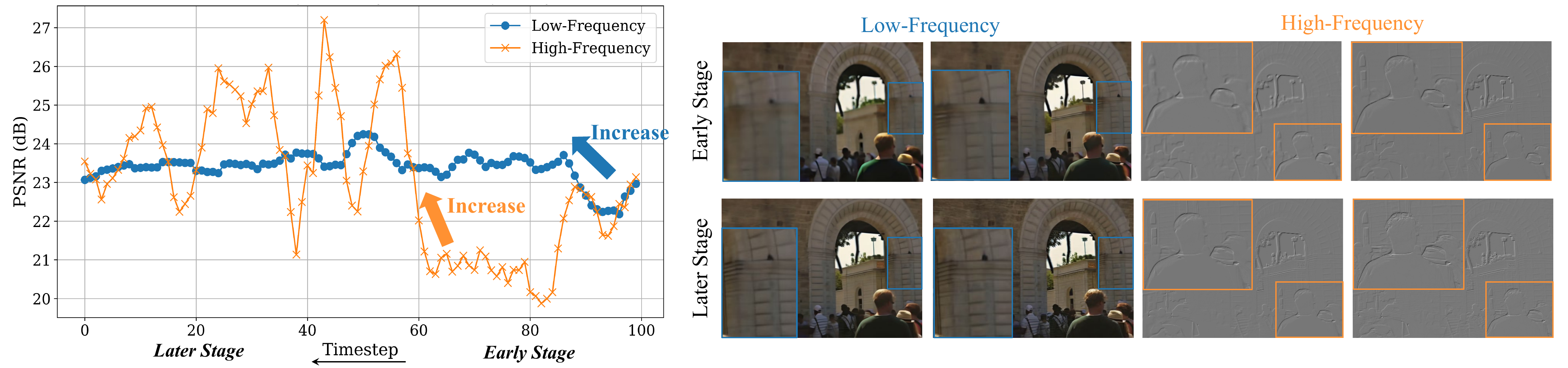}
    \caption{\textbf{Motivation of DF Loss.} \textbf{Left}: PSNR curves of low- and high-frequency components relative to ground truth across diffusion steps. The low-frequency PSNR increases during the early diffusion steps, while the high-frequency PSNR rises in the later diffusion steps. 
    \textbf{Right}: visual results of low- and high-frequency components at different diffusion stage. (\textbf{Zoom-in for best view})}
    \label{motivation:daf}
\end{figure*}

\subsection{Overview}
\label{subsec:overview}

\paragraph{Modules.}
The~\name~primarily includes four modules: VAE \cite{kingma2013auto}, text encoder \cite{radford2021learning, raffel2020exploring}, ControlNet \cite{zhang2023adding} and T2V model \cite{zhang2023i2vgen, yang2024cogvideox} with Local Information Enhancement Module (\textbf{LIEM}) to alleviate the artifacts (further analysis is provided in Sec.~\ref{subsec:LIEM}). 
As depicted in Figure~\ref{fig:overview}, the VAE encoder takes HR videos $X_H$ and LR videos $X_L$ as input to generate latent tensors $Z_H$ and $Z_L$, respectively. The text encoder is responsible for generating text embeddings $c_{text}$ to provide high-level information. ControlNet takes $Z_L$ and $c_{text}$ as input to guide the T2V model output. Finally, the T2V model $\phi_\theta$ with \textbf{LIEM} receives noisy input $Z_t = \alpha_t Z_H + \sigma_t \epsilon$ ($t$ denotes diffusion step, $\alpha_t$ and $\sigma_t$ are noise scheduler parameters), $c_{text}$ and the control signal from ControlNet $c_{l}$ to predict the velocity $v_t \equiv \alpha_t \epsilon - \sigma_t Z_H$ \cite{salimans2022progressive}. 

\vspace{-1em}
\paragraph{Losses.}
We utilize v-prediction objective in optimization:
\begin{equation}
    \mathcal{L}_{v} = \mathbb{E}[\| v_t - \phi_\theta(Z_t, c_{text}, c_{l}, t) \|_2^2].
\end{equation}
%
Given the strong generalization ability of T2V models, relying solely on the v-prediction objective for optimization may lead to restored outputs with low fidelity, an essential factor in video super-resolution tasks. 
To address this, we introduce Dynamic Frequency (\textbf{DF}) Loss, which adaptively adjusts the constraint on high- and low-frequency components of the predicted $\hat{X}_H$ across different diffusion steps. 
The overall optimization objective for~\name~is as follows:
\begin{equation}
    \mathcal{L}_{total} = \mathcal{L}_{v} + b(t)\mathcal{L}_{DF}(\hat{X}_H, X_H),
\end{equation}
where $b(t)= 1 - \frac{t}{t_{max}}$ is a weighting function ($t_{max}$ is set to 999) to balance $\mathcal{L}_{v}$ and $\mathcal{L}_{DF}$. 
With the proposed LIEM and DF loss,~\name~achieves high spatio-temporal quality, reduced artifacts and enhanced fidelity.

\subsection{Local Information Enhancement Module}
\label{subsec:LIEM}

\paragraph{Motivation.} 
%
Most T2V models primarily use a global attention mechanism~\cite{liu2021global}, which is well-suited to text-to-video tasks by capturing global information to generate complete videos from scratch. 
However, this approach may be suboptimal for real-world video super-resolution, where complex degradations occur and local details are crucial~\cite{kong2022residual}.
Relying solely on global attention mechanisms presents two drawbacks for real-world video super-resolution: 
$1$) It \textit{complicates degradation removal}, as it processes the entire degraded video at once (the first and second columns in Figure~\ref{motivation:liem} (right)).
$2$) It \textit{lacks local details}, resulting in blurry outputs (the third column in Figure~\ref{motivation:liem} (right)). 


\vspace{-1em}
\paragraph{Details of LIEM.}
To address the above issues, we propose a simple but effective approach: adding a \textit{Local Information Enhancement Module} (LIEM) before the global attention block to make T2V model pay more attention to local information. It can be expressed by:
\begin{equation}
    L(F_I) = Sigmoid(Conv_{3\times3}(Concat(AP(F_I), MP(F_I)))),
\end{equation}
\begin{equation}
    F_O = G(L(F_I) \cdot F_I) + F_I,
\end{equation}
where $AP(\cdot)$ and $MP(\cdot)$ denote average pooling and max pooling, respectively. $F_I$ and $F_O$ represent the input and output features, while $G(\cdot)$ and $L(\cdot)$ refer to the global attention block and LIEM. We adopt the local attention block in CBAM \cite{woo2018cbam} as LIEM for simplicity. 
Additional analysis on the impact of adding LIEM is provided in Sec.~\ref{sec:liem_ablation}.
%
%
Intuitively, as shown in the second row of Figure~\ref{motivation:liem} (left), 
incorporating LIEM enables the T2V model to address local region degradation first and then aggregate global features. 
This approach reduces the complexity of degradation removal and mitigates artifacts. 
Furthermore, the T2V model with LIEM produces clearer, more detailed results due to the enriched local information.
 
\subsection{Dynamic Frequency Loss}
\label{subsec:daf}

\begin{table*}[t!]
    \caption{Quantitative evaluations on diverse VSR benchmarks from synthetic (UDM10, REDS30, OpenVid30) and real-world (VideoLQ) sources. The best performance is highlighted in \textbf{bold}, and the second-best in \underline{underlined}. E$^*_{warp}$ refers to E$_{warp}$ ($\times10^{-3}$).} \label{tab:my_label}
    \centering
    \resizebox{\textwidth}{!}{
    \begin{tabular}{cc|cccc|cccccc}
    \hline
    \multirow{2}{*}{Datasets} & \multirow{2}{*}{Metrics} & Real-ESRGAN & DBVSR & RealBasicVSR & RealViformer & ResShift & StableSR & Upscale-A-Video & MGLDVSR & Ours \\ 
    ~ & ~ & ICCVW 2021 & ICCV 2021 & CVPR 2022 & ECCV 2024 & NeurIPS 2023 & IJCV 2024 & CVPR 2024 & ECCV 2024 & - \\ \hline \hline
    \multirow{5}{*}{UDM10} & PSNR$\uparrow$ & 22.41 & 19.65 & 23.64 & \textbf{24.00} & 22.90 & 23.50 & 21.29 & 23.74 & \underline{23.91} \\
    ~ & SSIM$\uparrow$ & 0.6476 & 0.4747 & 0.6842 & \underline{0.6896} & 0.5451 & 0.6599 & 0.5967 & 0.6826 & \textbf{0.7164} \\
    ~ & LPIPS$\downarrow$ & 0.2769 & 0.4566 & 0.2514 & 0.2325 & 0.4036 & 0.2785 & 0.3006 & \underline{0.2195} & \textbf{0.1885} \\
    ~ & DOVER$\uparrow$ & 0.4831 & 0.0959 & 0.5039 & 0.5055 & 0.3252 & 0.3490 & \underline{0.5309} & 0.4896 & \textbf{0.5422} \\
    ~ & E$^*_{warp}$ $\downarrow$ & 11.17 & 12.56 & 5.14 & 3.57 & 12.69 & 8.89 & \underline{2.83} & 6.03 & \textbf{2.68} \\ \hline
    \multirow{5}{*}{REDS30} & PSNR$\uparrow$ & 19.56 & 14.85 & \underline{20.85} & \textbf{20.86} & 19.93 & 20.32 & 19.71 & 20.57 & 20.29 \\
    ~ & SSIM$\uparrow$ & 0.4862 & 0.2941 & \textbf{0.5469} & 0.5377 & 0.4261 & 0.5043 & 0.4315 & 0.5113 & \underline{0.5411} \\
    ~ & LPIPS$\downarrow$ & 0.3376 & 0.5915 & 0.2899 & \underline{0.2597} & 0.4422 & 0.3857 & 0.3443 & \textbf{0.2240} & 0.2804 \\
    ~ & DOVER$\uparrow$ & 0.3182 & 0.0600 & 0.3483 & 0.3400 & 0.2221 & 0.2519 & 0.2857 & \underline{0.3857} & \textbf{0.4017} \\
    ~ & E$^*_{warp}$ $\downarrow$ & 19.1 & 18.00 & 8.32 & \textbf{6.06} & 17.40 & 22.14 & 15.65 & 12.28 & \underline{7.30} \\ \hline
    \multirow{5}{*}{OpenVid30} & PSNR$\uparrow$ & 24.62 & 21.14 & 24.63 & \textbf{26.21} & 24.29 & 24.91 & 24.41 & 24.73 & \underline{25.30} \\
    ~ & SSIM$\uparrow$ & 0.7778 & 0.5887 & 0.7759 & \underline{0.8080} & 0.6070 & 0.7633 & 0.7167 & 0.7686 & \textbf{0.8371} \\
    ~ & LPIPS$\downarrow$ & 0.1994 & 0.4207 & 0.2297 & \underline{0.1881} & 0.3902 & 0.2102 & 0.2479 & 0.2074 & \textbf{0.1011} \\
    ~ & DOVER$\uparrow$ & 0.6992 & 0.1819 & \underline{0.7345} & 0.7275 & 0.5435 & 0.6368 & 0.7201 & 0.7191 & \textbf{0.7393} \\
    ~ & E$^*_{warp}$ $\downarrow$ & 8.46 & 12.11 & 4.12 & \underline{2.52} & 9.78 & 8.87 & 4.72 & 4.82 & \textbf{1.82} \\ \hline \hline
    \multirow{4}{*}{VideoLQ} & ILNIQE$\downarrow$ & 27.95 & 27.19 & 26.29 & 26.11 & 25.92 & 29.97 & \underline{24.49} & \textbf{23.94} & 25.35 \\
    ~ & DOVER$\uparrow$ & 0.4967 & 0.3392 & \underline{0.5285} & 0.4804 & 0.4113 & 0.4775 & 0.4833 & 0.5319 & \textbf{0.5431} \\
    ~ & E$^*_{warp}$ $\downarrow$ & 8.00 & 7.75 & 6.52 & \textbf{5.10} & 8.33 & 9.26 & 10.89 & 7.82 & \underline{6.38} \\ \hline
    \end{tabular}}
\end{table*}

\paragraph{Motivation.}
The powerful generative capacity of diffusion models may compromise the fidelity in restored result~\cite{wu2024seesr, yu2024scaling}. 
In Figure~\ref{motivation:daf} (Right), an interesting pattern emerges when examining restored results at each diffusion step during inference. 
In the early stages, the model primarily reconstructs structure with low frequency, whereas in later stages, after the structure is largely complete, focus shifts to refining details with high frequency. 
To further illustrate this phenomenon, Figure~\ref{motivation:daf} (Left) presents PSNR curves of low- and high-frequency components against the ground truth across diffusion steps. 
The low-frequency PSNR rises in the early stages, while the high-frequency PSNR increases later, aligning with the visual results.


Fidelity can be divided into two types: 
$1$) Low-frequency fidelity, encompassing large structures and instances. 
2) High-frequency fidelity, including edges and textures, aligning with the characteristics of the denoising process. 
This raises a question: Can we design a loss function that exploits this characteristic to decouple fidelity and simplify optimization?
Specifically, we aim to guide the model to \textit{prioritize low-frequency components in the early stages}, shifting focus to \textit{high-frequency components later}.

\begin{figure}[t!]
    \centering
    \includegraphics[width=1\linewidth]{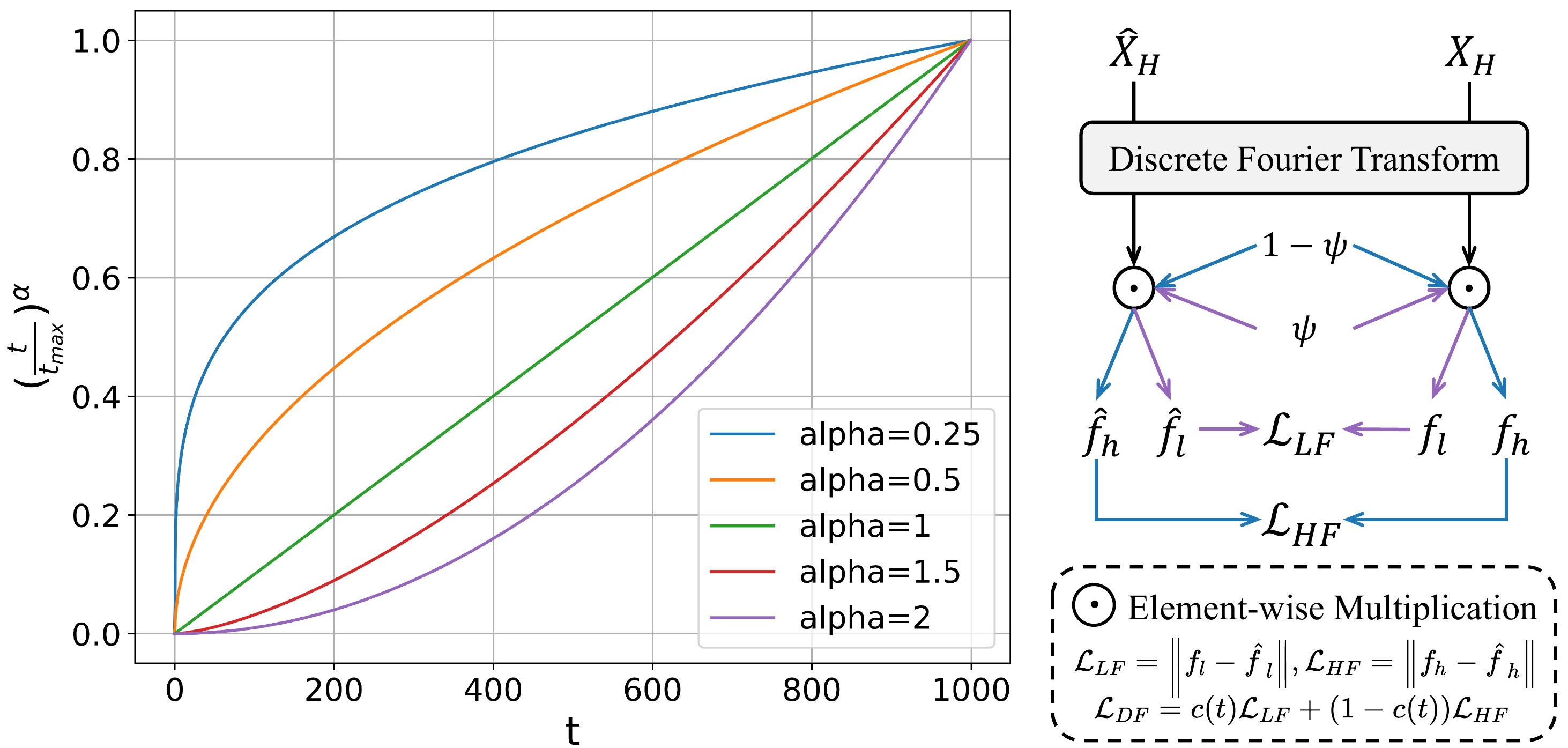}
    \caption{\textbf{Dynamic Frequency Loss.} \textbf{Left}: curves of weighting function $c(t)$ for different $\alpha$. \textbf{Right}: details of DF loss.}
    \label{fig:daf_method}
\end{figure}

\vspace{-1em}
\paragraph{Details of DF Loss.}
Here, we propose Dynamic Frequency Loss. 
Specifically, in each diffusion step $t$, we use the following equation to obtain the estimated $\hat{Z}_H$:
\begin{equation}
    \hat{Z}_H = \sigma_t^{-1}(\alpha_t\epsilon-\phi_\theta(Z_t, c_{text}, c_{l}, t)).
\end{equation}
Then, we use the decoder to convert the latent $\hat{Z}_H$ back to the pixel space, resulting in $\hat{X}_H$. After that, we apply Discrete Fourier Transform (DFT) to transform $\hat{X}_H$ into the frequency domain as shown in Figure~\ref{fig:daf_method}. 
We predefine a low-frequency pass filter $\psi$ to obtain the low- and high-frequency: 
\begin{equation}
    \hat{f}_l = \mathcal{F}(\hat{X}_H) \odot \psi, \hat{f}_h = \mathcal{F}(\hat{X}_H) \odot (1-\psi),
\end{equation}
where $\mathcal{F}(\cdot)$ is DFT, $\odot$ is element-wise multiplication. $\hat{f}_{l}$ and $\hat{f}_{h}$ denote the low and high frequency of $\hat{X}_H$. The proposed DF loss can be written as: 
\begin{equation}
    \mathcal{L}_{LF} = \| f_l - \hat{f}_l \|, \mathcal{L}_{HF} = \| f_h - \hat{f}_h \|,
\end{equation}
\begin{equation}
    \mathcal{L}_{DF} = c(t)\mathcal{L}_{LF} + (1-c(t))\mathcal{L}_{HF},
\end{equation}
where $f_l$ / $f_h$ stand for low- / high-frequency of $X_H$, respectively. $c(t) = (t/t_{max})^\alpha$ is the weighting function.

\section{Experiments}

\begin{figure*}[t!]
    \centering
    \includegraphics[width=1\linewidth]{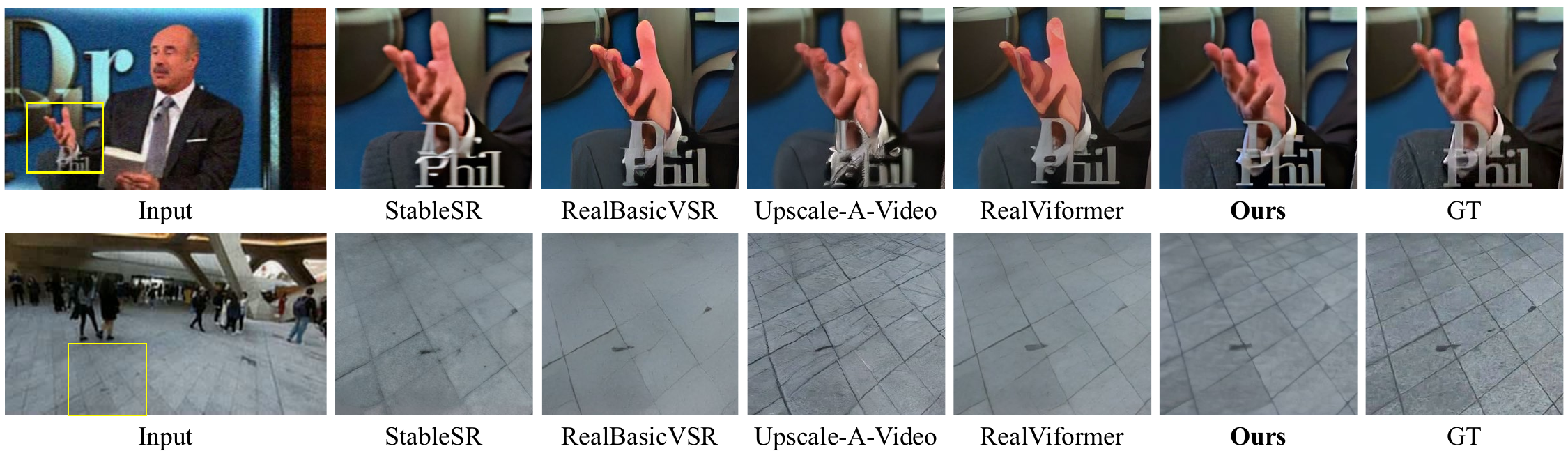}
    \caption{Qualitative comparisons on synthetic LR videos from OpenVid30 and REDS30\cite{nah2019ntire}. \textbf{(Zoom-in for best view)}}
    \label{fig:qualitive1}
\end{figure*}

\begin{figure*}[t!]
    \centering
    \includegraphics[width=1\linewidth]{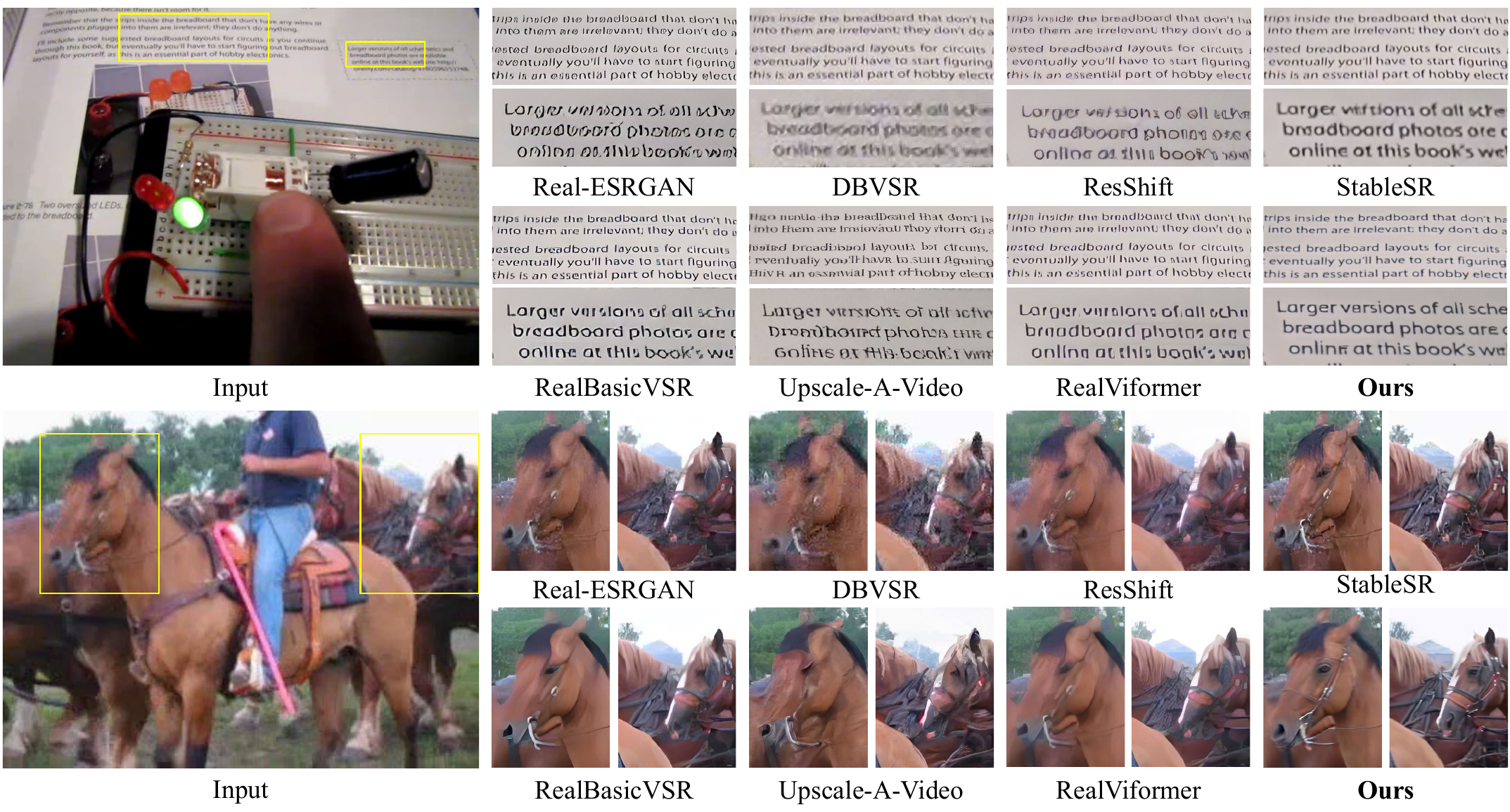}
    \caption{Qualitative comparisons on real-world test videos in VideoLQ \cite{chan2022investigating} dataset. \textbf{(Zoom-in for best view)}}
    \label{fig:qualitive2}
\end{figure*}

\begin{figure*}
    \centering
    \includegraphics[width=\linewidth]{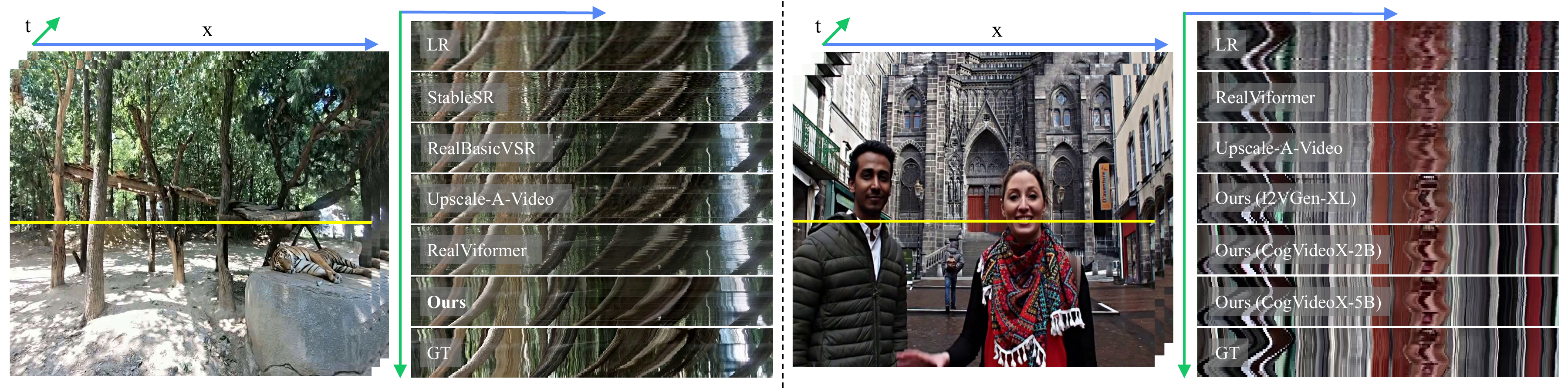}
    \caption{Qualitative comparisons on temporal consistency in REDS30 \cite{nah2019ntire} and OpenVid dataset. \textbf{(Zoom-in for best view)}}
    \label{fig:temp_consis}
\end{figure*}

\subsection{Datasets and Implementation}
\noindent
\textbf{Training Datasets.}
We train~\name~using the subset of OpenVid-1M~\cite{nan2024openvid}, containing $\sim$200K text-video pairs. The OpenVid-1M dataset is a high-quality video dataset consisting of over 1 million in-the-wild video clips with detailed captions, where the minimum resolution is $512$$\times$$512$ and the average length is 7.2 seconds. Utilizing this large-scale high-quality data for training further improves our model's restoration capacity for real-world VSR. More training dataset comparisons can be found in Table \ref{tab:dataset}.
We generate the LR-HR video pairs following the degradation strategy in Real-ESRGAN \cite{wang2021realesrgan}, combined with video compression operations, resulting in severe degradation similar to the approach used in RealBasicVSR \cite{chan2022investigating}.

\begin{table}[]
    \centering
    \caption{Training dataset comparison.}
    \resizebox{1\linewidth}{!}{
    \begin{tabular}{ccccc}
    \hline
       Method & Dataset & Size & $\#$Frames & Resolution \\ \hline
       UAV\cite{zhou2024upscale} & WebVid \cite{bain2021frozen} + YouHQ \cite{zhou2024upscale} & 335K+37K & - & 336$\times$596, 1080$\times$1920 \\
       RealViformer\cite{zhang2024realviformer} & REDS \cite{nah2019ntire} & 300K & 100 & 720$\times$1280 \\
       Ours & OpenVid \cite{nan2024openvid} & 200K & 32 & 720$\times$1280 \\ \hline
    \end{tabular}
    }
    \label{tab:dataset}
\end{table}

\noindent
\textbf{Testing Datasets.}
We evaluate our method on both synthetic and real-world datasets. 
As for synthetic testing datasets, we follow the same degradation pipeline in training to generate LR videos from HR ones to construct three synthetic datasets (\textit{i.e.}, UDM10 \cite{yi2019progressive}, REDS30 \cite{nah2019ntire}, and OpenVid30). The OpenVid30 is split from OpenVid-1M \cite{nan2024openvid} ensuring no overlap with the training dataset and comprises the first approximately 100 frames of 30 videos. For the real-world dataset, we choose VideoLQ \cite{chan2022investigating} which contains 50 videos, each with 100 frames.

\noindent
\textbf{Training Details.}
By default, we adopt I2VGen-XL~\cite{zhang2023i2vgen} as our T2V backbone.
For fast convergence, we initialize the model using the weights from VEnhancer~\cite{he2024venhancer}. We then train the ControlNet and inserted LIEM to adapt the T2V model for the real-world VSR task. Specifically, we train~\name~on $8$ NVIDIA A100-80G GPUs with $15$K iterations and a batch size of $8$. 
The training data is $720$$\times$$1280$ with $32$ frames. We use AdamW \cite{loshchilov2017decoupled} as the optimizer with a learning rate of 5e-5.

\noindent
\textbf{Evaluation Metrics.}
We adopt six metrics to evaluate the VSR outputs from several different perspectives: image fidelity (PSNR), perceptual similarity (SSIM \cite{wang2004image}, LPIPS \cite{zhang2018lpips}), quality (ILNIQE \cite{zhang2015ilniqe}), video clarity (DOVER \cite{wu2023dover}) and temporal consistency ($E^*_{warp}$ \cite{Lai2018warping, liu2024evalcrafter}). 
For synthetic datasets, we calculate PSNR, SSIM and LPIPS between the output and ground-truth frames, along with DOVER and flow warping error (\textit{i.e.}, $E^*_{warp}$) of output videos. For real-world dataset, because of no ground-truth videos, we use three non-reference metrics: ILNIQE, DOVER, and $E^*_{warp}$.

\subsection{Comparisons}
To verify the effectiveness of our approach, we compare~\name~with several state-of-the-art methods, including Real-ESRGAN \cite{wang2021realesrgan}, DBVSR \cite{pan2021deep}, RealBasicVSR \cite{chan2022investigating}, RealViformer \cite{zhang2024realviformer}, ResShift \cite{yue2024resshift}, StableSR \cite{wang2024exploiting}, and Upscale-A-Video \cite{zhou2024upscale}.

\noindent
\textbf{Quantitative Evaluation.}
As shown in Table \ref{tab:my_label}, we calculate five metrics on each synthetic benchmark. Our~\name~achieves the best scores in four out of these five metrics (SSIM, LPIPS, DOVER, and $E^*_{warp}$) on both UDM10 and OpenVid30 datasets, along with the second-best PSNR scores. This indicates that~\name~can generate realistic details with good fidelity and robust temporal consistency.
Moreover, we evaluate three non-reference metrics on a real-world dataset. On this dataset,~\name~achieves the best score in DOVER and the second-best scores in ILNIQE and $E^*_{warp}$. These results demonstrate that~\name~can effectively restore real-world videos with high spatial and temporal quality.
Additionally, our visual results on both real-world and synthetic datasets are preferred by human evaluators, as detailed in the User Study section (see Appendix).

\vspace{-0.3mm}
\noindent
\textbf{Qualitative Evaluation.}
To intuitively demonstrate the effectiveness of the proposed~\name, we present visual results on both synthetic and real-world datasets in Figure \ref{fig:qualitive1} and \ref{fig:qualitive2}, respectively. As shown, our~\name~generates the most realistic spatial details and exhibits the best degradation removal capability. Specifically, the first example in Figure \ref{fig:qualitive2} illustrates that ~\name~reconstructs the text structure most effectively, thanks to the T2V prior efficiently capturing temporal information, and the DF loss that improves the fidelity. Furthermore, the T2V model has a strong spatial prior, which helps generate more realistic details and structures, such as the human hand in Figure \ref{fig:qualitive1} and the horse shape and fur in Figure \ref{fig:qualitive2}.

We also compare the temporal consistency in Figure \ref{fig:temp_consis}. As observed in the left of Figure \ref{fig:temp_consis}, StableSR demonstrates the most temporal inconsistency, primarily because it is originally designed for image super-resolution. Although RealBasicVSR, Upscale-A-Video, and RealViformer incorporate optical flow maps to enhance temporal consistency, they still face challenges in generating consistent results under complex degraded video conditions, as the optical flow maps may not always be accurate. In contrast, our proposed~\name~achieves the best temporal consistency, thanks to the powerful temporal prior inherent in the T2V model, which effectively helps reconstruct temporal information \textit{even without the use of optical flow maps.}


\subsection{Ablation Study}

\begin{figure*}
    \centering
    \includegraphics[width=\linewidth]{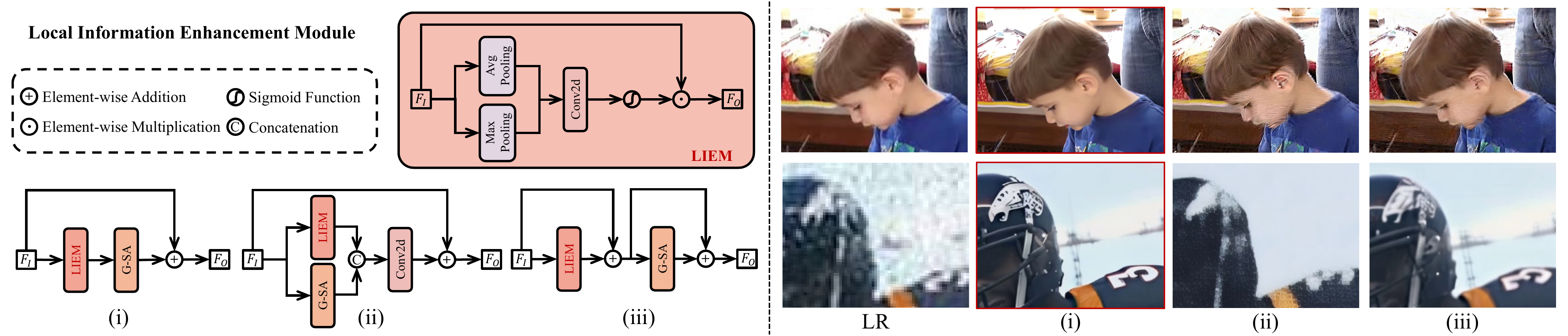}
    \caption{Ablation study about LIEM. \textbf{Left:} illustration of different insertion positions of LIEM and the structure of LIEM. \textbf{Right:} visual comparison on real-world and synthetic videos with different LIEM positions.}
    \label{fig:liem_ablation}
\end{figure*}

\begin{table}[t]
    \centering
    \caption{Ablation of LIEM position.}
    \resizebox{\linewidth}{!}{
    \begin{tabular}{c|cc|ccc}
    \hline
       Position & Spa-Local & Temp-Local & PSNR$\uparrow$ & LPIPS$\downarrow$ & $E_{warp}^*\downarrow$ \\ \hline
        \multirow{4}{*}{(i)} & ~ & ~ & 23.14 & 0.2015 & 2.83 \\
        ~ & $\checkmark$ & ~ & 23.61 & 0.2013 & 2.82 \\
        ~ & ~ & $\checkmark$ & 23.65 & \underline{0.1945} & 2.92 \\ \cline{2-3}
        ~ & \multirow{3}{*}{$\checkmark$} & \multirow{3}{*}{$\checkmark$} & \cellcolor{lightgray}\underline{23.69} & \cellcolor{lightgray}\textbf{0.1943} & \cellcolor{lightgray}\underline{2.74} \\ \cline{1-1}
        (ii) & ~ & ~ & 23.27 & 0.2363 & 3.57 \\
        (iii) & ~ & ~ & \textbf{24.51} & 0.2094 & \textbf{1.99} \\ \hline
    \end{tabular}}
    \label{tab:position_ablation}
\end{table}

\noindent
\textbf{Local Information Enhancement Module.}
\label{sec:liem_ablation}
We primarily investigate the impact of introducing LIEM in different ways. First, we find that adding LIEM on both spatial and temporal blocks achieves the best results as shown in Table \ref{tab:position_ablation}. Second, we consider three connection types as shown in Figure \ref{fig:liem_ablation} (Left). From visual results in Figure \ref{fig:liem_ablation} (Right) and quantitative results in Table \ref{tab:position_ablation}, we find that position (i) achieves the best results. This phenomenon can be attributed to the fact that, with most weights frozen to preserve the prior, the newly added blocks can influence the model's mapping process. However, the impact at positions (ii) and (iii) is too large, making it difficult for the model to fine-tune and adapt to this change, resulting in poor performance.

\begin{table}[t]
    \centering
    \caption{Ablation of different variants of DF loss.}
    \begin{tabular}{cc|ccc}
    \hline
       Seperate & Type & PSNR$\uparrow$ & LPIPS$\downarrow$ & $E_{warp}^*\downarrow$ \\ \hline
       \multicolumn{2}{c|}{w/o Frequency Loss} & 23.69 & 0.1943 & 2.74 \\ \hline
        - & - & \underline{23.72} & \underline{0.1941} & \underline{2.71} \\
        $\checkmark$ & Inverse & 23.67 & 0.1945 & 2.83 \\
        $\checkmark$ & Direct & \cellcolor{lightgray}\textbf{23.85} & \cellcolor{lightgray}\textbf{0.1903} & \cellcolor{lightgray}\textbf{2.69} \\ \hline
    \end{tabular}
    \label{tab:daf_ablation1}
\end{table}

\begin{table}[t]
    \centering
    \caption{Ablation of $b(t)$ and $\alpha$ in $c(t)$.}
    \begin{tabular}{c|c|ccc}
    \hline
       b(t) & $\alpha$ & PSNR$\uparrow$ & LPIPS$\downarrow$ & $E_{warp}^*\downarrow$ \\ \hline
       \multirow{5}{*}{Linear} & 0.25 & 23.76 & 0.2030 & 2.72 \\
        ~ & 0.5 & 23.71 & 0.2010 & 2.75 \\
        ~ & 1 & \underline{23.85} & \underline{0.1903} & \underline{2.69} \\
        ~ & 1.5 & 23.53 & 0.1928 & 2.81 \\ \cline{2-2}
        ~ & \multirow{2}{*}{2} & \cellcolor{lightgray}\textbf{23.91} & \cellcolor{lightgray}\textbf{0.1885} & \cellcolor{lightgray}\textbf{2.61} \\ \cline{1-1}
        Exponential & ~ & 23.68 & 0.1990 & 2.78 \\
        \hline
    \end{tabular}
    \label{tab:daf_ablation2}
\end{table}



\begin{figure}
    \centering
    \includegraphics[width=1\linewidth]{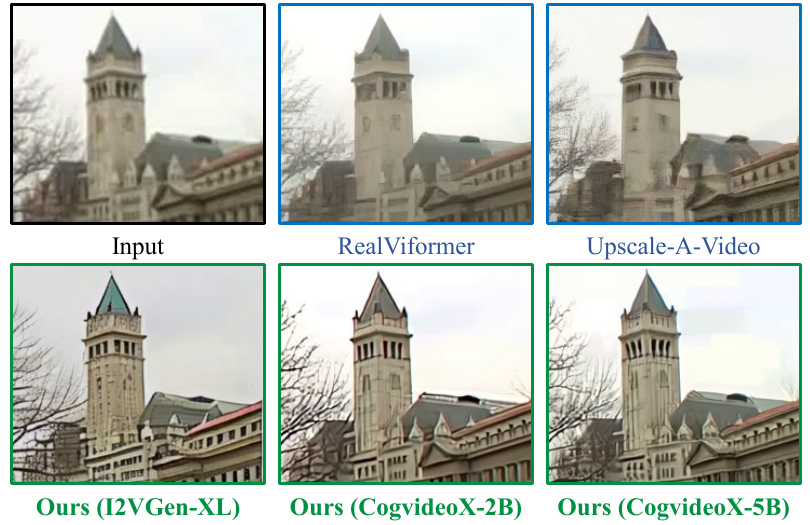}
    \caption{Illustration on scaling up with larger t2v models on a real-world low-quality video. \textbf{(Zoom-in for best view)}}
    \label{fig:ablation3}
\end{figure}

\noindent
\textbf{Dynamic Frequency Loss.}
\label{ablation:daf}
First, we investigate the impact of different variants of frequency loss. As shown in Table~\ref{tab:daf_ablation1}, ``Separate" indicates whether the frequency components are separated into high and low frequency, constraining them individually. 
``Type" refers to the specific definition of the DF loss: if set to ``inverse," a higher weight is given to high frequencies in the early stages and a lower weight to low frequencies; if set to ``direct", a higher weight is given to low frequencies initially and a lower weight to high frequencies, which is matching the analysis in Sec.~\ref{subsec:daf}. As observed, separating the frequency components and prioritizing low-frequency reconstruction early on yield the best perceptual quality while maintaining high fidelity. Second, we explore the optimal settings for $b(t)$ and $\alpha$ in $c(t)$. As shown in Table~\ref{tab:daf_ablation2}, using a linear form for $b(t)$ with $\alpha=2$ for $c(t)$ yields the best results. Therefore, we adopt this DF loss configuration for training our model and comparing it with other state-of-the-art methods.


\begin{table}[t]
    \centering
    \caption{Effectiveness of T2V diffusion prior for real-world VSR.}
    \resizebox{\linewidth}{!}{
    \begin{tabular}{c|cc|ccc}
    \hline
       \multirow{2}{*}{Metrics} & \multirow{2}{*}{UAV} & \multirow{2}{*}{RealViformer} & \multicolumn{3}{c}{Ours} \\
       ~ & ~ & ~ & I2VGen-XL & CogX-2B & CogX-5B \\ \hline
        PSNR$\uparrow$ & 22.46& 22.90& 21.46& \underline{23.18}& \textbf{23.60}\\
        SSIM$\uparrow$ & 0.6552& 0.6944& 0.6715& \underline{0.7112}& \textbf{0.7400}\\
        LPIPS$\downarrow$ & 0.2035& 0.1823& 0.1779& \underline{0.1571}& \textbf{0.1314}\\
        DOVER$\uparrow$ & 0.6609& 0.4286& \underline{0.7267}& 0.6955& \textbf{0.7350}\\
        $E_{warp}^*\downarrow$ & 5.424& 4.75& 5.529& \textbf{3.68} & \underline{4.56}\\ \hline
    \end{tabular}}
    \label{tab:t2v prior}
\end{table}
\vspace{-1mm}

\noindent
\textbf{Scaling up with Larger T2V Models.}
To further validate the effectiveness of T2V diffusion priors for real-world VSR, we replace I2VGen-XL with larger 
DiT-based \cite{peebles2023scalable} T2V models (\textit{i.e.,} CogVideoX~\cite{cogvideox5b,yang2024cogvideox}), and evaluate results both quantitatively and qualitatively. 
Since CogVideoX only supports inputs at 480$\times$720 resolution, we created a new test set by cropping 10 videos from OpenVid-1M \cite{nan2024openvid} to this size. 
As shown in Table~\ref{tab:t2v prior}, the powerful CogVideoX models yield consistent improvements across all metrics. 
%
%
Notably, SSIM improves from 0.6944 to 0.7400, and DOVER increases from 0.6609 to 0.7350, marking a substantial enhancement in visual quality. 
The robust spatio-temporal priors in CogVideoX enable realistic details and clear building structures (Figure~\ref{fig:ablation3}), while maintaining high temporal consistency (Figure~\ref{fig:temp_consis} Right).
Inspired by scaling law \cite{henighan2020scaling, kaplan2020scaling} and our findings, we believe larger, more powerful T2V models will further advance VSR tasks.


\section{Conclusion}
In this paper, we present~\name, a real-world VSR framework that leverages T2V diffusion prior to restore videos with fewer artifacts, higher spatial fidelity, and stronger temporal consistency. Specifically, we introduce a Local Information Enhancement Module into the original T2V backbone to improve its ability to handle degradations and reconstruct fine details. Additionally, we propose a Dynamic Frequency Loss that guides the model to focus on restoring different frequency components at each diffusion step, thereby enhancing fidelity. Furthermore, we demonstrate that a powerful T2V model can effectively generate high-quality results in both spatial and temporal dimensions. Extensive experiments show that~\name~achieves superior performance in both spatial and temporal quality. We hope our work lays a solid foundation for applying T2V models in real-world VSR and inspires future advancements in the field.
{
    \small
    \bibliographystyle{ieeenat_fullname}
    \bibliography{main}

\begin{thebibliography}{75}
\providecommand{\natexlab}[1]{#1}
\providecommand{\url}[1]{\texttt{#1}}
\expandafter\ifx\csname urlstyle\endcsname\relax
  \providecommand{\doi}[1]{doi: #1}\else
  \providecommand{\doi}{doi: \begingroup \urlstyle{rm}\Url}\fi

\bibitem[cog(2024)]{cogvideox5b}
Cogvideox-5b, 2024.
\newblock \url{https://huggingface.co/THUDM/CogVideoX-5b}.

\bibitem[Bain et~al.(2021)Bain, Nagrani, Varol, and Zisserman]{bain2021frozen}
Max Bain, Arsha Nagrani, G{\"u}l Varol, and Andrew Zisserman.
\newblock Frozen in time: A joint video and image encoder for end-to-end retrieval.
\newblock In \emph{ICCV}, pages 1728--1738, 2021.

\bibitem[Bao et~al.(2024)Bao, Xiang, Yue, He, Zhu, Zheng, Zhao, Liu, Wang, and Zhu]{bao2024vidu}
Fan Bao, Chendong Xiang, Gang Yue, Guande He, Hongzhou Zhu, Kaiwen Zheng, Min Zhao, Shilong Liu, Yaole Wang, and Jun Zhu.
\newblock Vidu: a highly consistent, dynamic and skilled text-to-video generator with diffusion models.
\newblock \emph{arXiv preprint arXiv:2405.04233}, 2024.

\bibitem[Blattmann et~al.(2023{\natexlab{a}})Blattmann, Dockhorn, Kulal, Mendelevitch, Kilian, Lorenz, Levi, English, Voleti, Letts, et~al.]{blattmann2023stable}
Andreas Blattmann, Tim Dockhorn, Sumith Kulal, Daniel Mendelevitch, Maciej Kilian, Dominik Lorenz, Yam Levi, Zion English, Vikram Voleti, Adam Letts, et~al.
\newblock Stable video diffusion: Scaling latent video diffusion models to large datasets.
\newblock \emph{arXiv preprint arXiv:2311.15127}, 2023{\natexlab{a}}.

\bibitem[Blattmann et~al.(2023{\natexlab{b}})Blattmann, Rombach, Ling, Dockhorn, Kim, Fidler, and Kreis]{blattmann2023align}
Andreas Blattmann, Robin Rombach, Huan Ling, Tim Dockhorn, Seung~Wook Kim, Sanja Fidler, and Karsten Kreis.
\newblock Align your latents: High-resolution video synthesis with latent diffusion models.
\newblock In \emph{CVPR}, pages 22563--22575, 2023{\natexlab{b}}.

\bibitem[Blau and Michaeli(2018)]{blau2018perception}
Yochai Blau and Tomer Michaeli.
\newblock The perception-distortion tradeoff.
\newblock In \emph{Proceedings of the IEEE conference on computer vision and pattern recognition}, pages 6228--6237, 2018.

\bibitem[Brooks et~al.(2024)Brooks, Peebles, Holmes, DePue, Guo, Jing, Schnurr, Taylor, Luhman, Luhman, Ng, Wang, and Ramesh]{videoworldsimulators2024}
Tim Brooks, Bill Peebles, Connor Holmes, Will DePue, Yufei Guo, Li Jing, David Schnurr, Joe Taylor, Troy Luhman, Eric Luhman, Clarence Ng, Ricky Wang, and Aditya Ramesh.
\newblock Video generation models as world simulators.
\newblock 2024.

\bibitem[Caballero et~al.(2017)Caballero, Ledig, Aitken, Acosta, Totz, Wang, and Shi]{caballero2017real}
Jose Caballero, Christian Ledig, Andrew Aitken, Alejandro Acosta, Johannes Totz, Zehan Wang, and Wenzhe Shi.
\newblock Real-time video super-resolution with spatio-temporal networks and motion compensation.
\newblock In \emph{CVPR}, pages 4778--4787, 2017.

\bibitem[Chan et~al.(2021)Chan, Wang, Yu, Dong, and Loy]{chan2021basicvsr}
Kelvin~CK Chan, Xintao Wang, Ke Yu, Chao Dong, and Chen~Change Loy.
\newblock Basicvsr: The search for essential components in video super-resolution and beyond.
\newblock In \emph{CVPR}, pages 4947--4956, 2021.

\bibitem[Chan et~al.(2022{\natexlab{a}})Chan, Zhou, Xu, and Loy]{chan2022basicvsr++}
Kelvin~CK Chan, Shangchen Zhou, Xiangyu Xu, and Chen~Change Loy.
\newblock Basicvsr++: Improving video super-resolution with enhanced propagation and alignment.
\newblock In \emph{CVPR}, pages 5972--5981, 2022{\natexlab{a}}.

\bibitem[Chan et~al.(2022{\natexlab{b}})Chan, Zhou, Xu, and Loy]{chan2022investigating}
Kelvin~CK Chan, Shangchen Zhou, Xiangyu Xu, and Chen~Change Loy.
\newblock Investigating tradeoffs in real-world video super-resolution.
\newblock In \emph{CVPR}, pages 5962--5971, 2022{\natexlab{b}}.

\bibitem[Chen et~al.(2024{\natexlab{a}})Chen, Xu, Ren, Cong, He, Xie, Sinha, Luo, Xiang, and Perez-Rua]{chen2024gentron}
Shoufa Chen, Mengmeng Xu, Jiawei Ren, Yuren Cong, Sen He, Yanping Xie, Animesh Sinha, Ping Luo, Tao Xiang, and Juan-Manuel Perez-Rua.
\newblock Gentron: Diffusion transformers for image and video generation.
\newblock In \emph{Proceedings of the IEEE/CVF Conference on Computer Vision and Pattern Recognition}, pages 6441--6451, 2024{\natexlab{a}}.

\bibitem[Chen et~al.(2024{\natexlab{b}})Chen, Siarohin, Menapace, Deyneka, Chao, Jeon, Fang, Lee, Ren, Yang, et~al.]{chen2024panda}
Tsai-Shien Chen, Aliaksandr Siarohin, Willi Menapace, Ekaterina Deyneka, Hsiang-wei Chao, Byung~Eun Jeon, Yuwei Fang, Hsin-Ying Lee, Jian Ren, Ming-Hsuan Yang, et~al.
\newblock Panda-70m: Captioning 70m videos with multiple cross-modality teachers.
\newblock In \emph{CVPR}, pages 13320--13331, 2024{\natexlab{b}}.

\bibitem[Chen et~al.(2024{\natexlab{c}})Chen, Long, Qiu, Yao, Zhou, Luo, and Mei]{chen2024learning}
Zhikai Chen, Fuchen Long, Zhaofan Qiu, Ting Yao, Wengang Zhou, Jiebo Luo, and Tao Mei.
\newblock Learning spatial adaptation and temporal coherence in diffusion models for video super-resolution.
\newblock In \emph{CVPR}, pages 9232--9241, 2024{\natexlab{c}}.

\bibitem[Fuoli et~al.(2019)Fuoli, Gu, and Timofte]{fuoli2019efficient}
Dario Fuoli, Shuhang Gu, and Radu Timofte.
\newblock Efficient video super-resolution through recurrent latent space propagation.
\newblock In \emph{2019 IEEE/CVF International Conference on Computer Vision Workshop (ICCVW)}, pages 3476--3485. IEEE, 2019.

\bibitem[Haris et~al.(2019)Haris, Shakhnarovich, and Ukita]{haris2019recurrent}
Muhammad Haris, Gregory Shakhnarovich, and Norimichi Ukita.
\newblock Recurrent back-projection network for video super-resolution.
\newblock In \emph{CVPR}, pages 3897--3906, 2019.

\bibitem[He et~al.(2024)He, Xue, Liu, Lin, Gao, Lin, Qiao, Ouyang, and Liu]{he2024venhancer}
Jingwen He, Tianfan Xue, Dongyang Liu, Xinqi Lin, Peng Gao, Dahua Lin, Yu Qiao, Wanli Ouyang, and Ziwei Liu.
\newblock Venhancer: Generative space-time enhancement for video generation.
\newblock \emph{arXiv preprint arXiv:2407.07667}, 2024.

\bibitem[Henighan et~al.(2020)Henighan, Kaplan, Katz, Chen, Hesse, Jackson, Jun, Brown, Dhariwal, Gray, et~al.]{henighan2020scaling}
Tom Henighan, Jared Kaplan, Mor Katz, Mark Chen, Christopher Hesse, Jacob Jackson, Heewoo Jun, Tom~B Brown, Prafulla Dhariwal, Scott Gray, et~al.
\newblock Scaling laws for autoregressive generative modeling.
\newblock \emph{arXiv preprint arXiv:2010.14701}, 2020.

\bibitem[Ho et~al.(2022)Ho, Chan, Saharia, Whang, Gao, Gritsenko, Kingma, Poole, Norouzi, Fleet, et~al.]{ho2022imagen}
Jonathan Ho, William Chan, Chitwan Saharia, Jay Whang, Ruiqi Gao, Alexey Gritsenko, Diederik~P Kingma, Ben Poole, Mohammad Norouzi, David~J Fleet, et~al.
\newblock Imagen video: High definition video generation with diffusion models.
\newblock \emph{arXiv preprint arXiv:2210.02303}, 2022.

\bibitem[Huang et~al.(2017)Huang, Wang, and Wang]{huang2017video}
Yan Huang, Wei Wang, and Liang Wang.
\newblock Video super-resolution via bidirectional recurrent convolutional networks.
\newblock \emph{IEEE TPAMI}, 40\penalty0 (4):\penalty0 1015--1028, 2017.

\bibitem[Isobe et~al.(2020)Isobe, Jia, Gu, Li, Wang, and Tian]{isobe2020video}
Takashi Isobe, Xu Jia, Shuhang Gu, Songjiang Li, Shengjin Wang, and Qi Tian.
\newblock Video super-resolution with recurrent structure-detail network.
\newblock In \emph{ECCV}, pages 645--660. Springer, 2020.

\bibitem[Jo et~al.(2018)Jo, Oh, Kang, and Kim]{jo2018deep}
Younghyun Jo, Seoung~Wug Oh, Jaeyeon Kang, and Seon~Joo Kim.
\newblock Deep video super-resolution network using dynamic upsampling filters without explicit motion compensation.
\newblock In \emph{CVPR}, pages 3224--3232, 2018.

\bibitem[Kaplan et~al.(2020)Kaplan, McCandlish, Henighan, Brown, Chess, Child, Gray, Radford, Wu, and Amodei]{kaplan2020scaling}
Jared Kaplan, Sam McCandlish, Tom Henighan, Tom~B Brown, Benjamin Chess, Rewon Child, Scott Gray, Alec Radford, Jeffrey Wu, and Dario Amodei.
\newblock Scaling laws for neural language models.
\newblock \emph{arXiv preprint arXiv:2001.08361}, 2020.

\bibitem[Kingma(2013)]{kingma2013auto}
Diederik~P Kingma.
\newblock Auto-encoding variational bayes.
\newblock \emph{arXiv preprint arXiv:1312.6114}, 2013.

\bibitem[Kong et~al.(2022)Kong, Li, Liu, Liu, He, Bai, Chen, and Fu]{kong2022residual}
Fangyuan Kong, Mingxi Li, Songwei Liu, Ding Liu, Jingwen He, Yang Bai, Fangmin Chen, and Lean Fu.
\newblock Residual local feature network for efficient super-resolution.
\newblock In \emph{CVPR}, pages 766--776, 2022.

\bibitem[Lai et~al.(2018)Lai, Huang, Wang, Shechtman, Yumer, and Yang]{Lai2018warping}
Wei-Sheng Lai, Jia-Bin Huang, Oliver Wang, Eli Shechtman, Ersin Yumer, and Ming-Hsuan Yang.
\newblock Learning blind video temporal consistency.
\newblock In \emph{ECCV}, 2018.

\bibitem[Li et~al.(2020)Li, Tao, Guo, Qi, Lu, and Jia]{li2020mucan}
Wenbo Li, Xin Tao, Taian Guo, Lu Qi, Jiangbo Lu, and Jiaya Jia.
\newblock Mucan: Multi-correspondence aggregation network for video super-resolution.
\newblock In \emph{ECCV}, pages 335--351. Springer, 2020.

\bibitem[Liang et~al.(2022)Liang, Fan, Xiang, Ranjan, Ilg, Green, Cao, Zhang, Timofte, and Gool]{liang2022recurrent}
Jingyun Liang, Yuchen Fan, Xiaoyu Xiang, Rakesh Ranjan, Eddy Ilg, Simon Green, Jiezhang Cao, Kai Zhang, Radu Timofte, and Luc~V Gool.
\newblock Recurrent video restoration transformer with guided deformable attention.
\newblock \emph{NeurIPS}, 35:\penalty0 378--393, 2022.

\bibitem[Liang et~al.(2024)Liang, Cao, Fan, Zhang, Ranjan, Li, Timofte, and Van~Gool]{liang2024vrt}
Jingyun Liang, Jiezhang Cao, Yuchen Fan, Kai Zhang, Rakesh Ranjan, Yawei Li, Radu Timofte, and Luc Van~Gool.
\newblock Vrt: A video restoration transformer.
\newblock \emph{IEEE TIP}, 2024.

\bibitem[Lin et~al.(2023)Lin, He, Chen, Lyu, Dai, Yu, Ouyang, Qiao, and Dong]{lin2023diffbir}
Xinqi Lin, Jingwen He, Ziyan Chen, Zhaoyang Lyu, Bo Dai, Fanghua Yu, Wanli Ouyang, Yu Qiao, and Chao Dong.
\newblock Diffbir: Towards blind image restoration with generative diffusion prior.
\newblock \emph{arXiv preprint arXiv:2308.15070}, 2023.

\bibitem[Liu et~al.(2021)Liu, Shao, and Hoffmann]{liu2021global}
Yichao Liu, Zongru Shao, and Nico Hoffmann.
\newblock Global attention mechanism: Retain information to enhance channel-spatial interactions.
\newblock \emph{arXiv preprint arXiv:2112.05561}, 2021.

\bibitem[Liu et~al.(2024)Liu, Cun, Liu, Wang, Zhang, Chen, Liu, Zeng, Chan, and Shan]{liu2024evalcrafter}
Yaofang Liu, Xiaodong Cun, Xuebo Liu, Xintao Wang, Yong Zhang, Haoxin Chen, Yang Liu, Tieyong Zeng, Raymond Chan, and Ying Shan.
\newblock Evalcrafter: Benchmarking and evaluating large video generation models.
\newblock In \emph{CVPR}, pages 22139--22149, 2024.

\bibitem[Loshchilov(2017)]{loshchilov2017decoupled}
I Loshchilov.
\newblock Decoupled weight decay regularization.
\newblock \emph{arXiv preprint arXiv:1711.05101}, 2017.

\bibitem[Mikolov et~al.(2010)Mikolov, Karafi{\'a}t, Burget, Cernock{\`y}, and Khudanpur]{mikolov2010recurrent}
Tomas Mikolov, Martin Karafi{\'a}t, Lukas Burget, Jan Cernock{\`y}, and Sanjeev Khudanpur.
\newblock Recurrent neural network based language model.
\newblock In \emph{Interspeech}, pages 1045--1048. Makuhari, 2010.

\bibitem[Nah et~al.(2019)Nah, Baik, Hong, Moon, Son, Timofte, and Mu~Lee]{nah2019ntire}
Seungjun Nah, Sungyong Baik, Seokil Hong, Gyeongsik Moon, Sanghyun Son, Radu Timofte, and Kyoung Mu~Lee.
\newblock Ntire 2019 challenge on video deblurring and super-resolution: Dataset and study.
\newblock In \emph{CVPRW}, pages 0--0, 2019.

\bibitem[Nan et~al.(2024)Nan, Xie, Zhou, Fan, Yang, Chen, Li, Yang, and Tai]{nan2024openvid}
Kepan Nan, Rui Xie, Penghao Zhou, Tiehan Fan, Zhenheng Yang, Zhijie Chen, Xiang Li, Jian Yang, and Ying Tai.
\newblock Openvid-1m: A large-scale high-quality dataset for text-to-video generation.
\newblock \emph{arXiv preprint arXiv:2407.02371}, 2024.

\bibitem[OpenAI(2024)]{sora}
OpenAI.
\newblock Sora, 2024.
\newblock \url{https://openai.com/index/sora}.

\bibitem[Pan et~al.(2021)Pan, Bai, Dong, Zhang, and Tang]{pan2021deep}
Jinshan Pan, Haoran Bai, Jiangxin Dong, Jiawei Zhang, and Jinhui Tang.
\newblock Deep blind video super-resolution.
\newblock In \emph{ICCV}, pages 4811--4820, 2021.

\bibitem[Peebles and Xie(2023)]{peebles2023scalable}
William Peebles and Saining Xie.
\newblock Scalable diffusion models with transformers.
\newblock In \emph{ICCV}, pages 4195--4205, 2023.

\bibitem[Polyak et~al.(2024)Polyak, Zohar, Brown, Tjandra, Sinha, Lee, Vyas, Shi, Ma, Chuang, et~al.]{polyak2024movie}
Adam Polyak, Amit Zohar, Andrew Brown, Andros Tjandra, Animesh Sinha, Ann Lee, Apoorv Vyas, Bowen Shi, Chih-Yao Ma, Ching-Yao Chuang, et~al.
\newblock Movie gen: A cast of media foundation models.
\newblock \emph{arXiv preprint arXiv:2410.13720}, 2024.

\bibitem[Radford et~al.(2021)Radford, Kim, Hallacy, Ramesh, Goh, Agarwal, Sastry, Askell, Mishkin, Clark, et~al.]{radford2021learning}
Alec Radford, Jong~Wook Kim, Chris Hallacy, Aditya Ramesh, Gabriel Goh, Sandhini Agarwal, Girish Sastry, Amanda Askell, Pamela Mishkin, Jack Clark, et~al.
\newblock Learning transferable visual models from natural language supervision.
\newblock In \emph{International conference on machine learning}, pages 8748--8763. PMLR, 2021.

\bibitem[Raffel et~al.(2020)Raffel, Shazeer, Roberts, Lee, Narang, Matena, Zhou, Li, and Liu]{raffel2020exploring}
Colin Raffel, Noam Shazeer, Adam Roberts, Katherine Lee, Sharan Narang, Michael Matena, Yanqi Zhou, Wei Li, and Peter~J Liu.
\newblock Exploring the limits of transfer learning with a unified text-to-text transformer.
\newblock \emph{Journal of machine learning research}, 21\penalty0 (140):\penalty0 1--67, 2020.

\bibitem[Rombach et~al.(2022)Rombach, Blattmann, Lorenz, Esser, and Ommer]{rombach2022high}
Robin Rombach, Andreas Blattmann, Dominik Lorenz, Patrick Esser, and Bj{\"o}rn Ommer.
\newblock High-resolution image synthesis with latent diffusion models.
\newblock In \emph{CVPR}, pages 10684--10695, 2022.

\bibitem[Sajjadi et~al.(2018)Sajjadi, Vemulapalli, and Brown]{sajjadi2018frame}
Mehdi~SM Sajjadi, Raviteja Vemulapalli, and Matthew Brown.
\newblock Frame-recurrent video super-resolution.
\newblock In \emph{CVPR}, pages 6626--6634, 2018.

\bibitem[Salimans and Ho(2022)]{salimans2022progressive}
Tim Salimans and Jonathan Ho.
\newblock Progressive distillation for fast sampling of diffusion models.
\newblock \emph{arXiv preprint arXiv:2202.00512}, 2022.

\bibitem[Shi et~al.(2022)Shi, Gu, Xie, Wang, Yang, and Dong]{shi2022rethinking}
Shuwei Shi, Jinjin Gu, Liangbin Xie, Xintao Wang, Yujiu Yang, and Chao Dong.
\newblock Rethinking alignment in video super-resolution transformers.
\newblock \emph{NeurIPS}, 35:\penalty0 36081--36093, 2022.

\bibitem[Singer et~al.(2022)Singer, Polyak, Hayes, Yin, An, Zhang, Hu, Yang, Ashual, Gafni, et~al.]{singer2022make}
Uriel Singer, Adam Polyak, Thomas Hayes, Xi Yin, Jie An, Songyang Zhang, Qiyuan Hu, Harry Yang, Oron Ashual, Oran Gafni, et~al.
\newblock Make-a-video: Text-to-video generation without text-video data.
\newblock \emph{arXiv preprint arXiv:2209.14792}, 2022.

\bibitem[Wang et~al.(2024)Wang, Yue, Zhou, Chan, and Loy]{wang2024exploiting}
Jianyi Wang, Zongsheng Yue, Shangchen Zhou, Kelvin~CK Chan, and Chen~Change Loy.
\newblock Exploiting diffusion prior for real-world image super-resolution.
\newblock \emph{IJCV}, pages 1--21, 2024.

\bibitem[Wang and Yang(2024)]{wang2024vidprom}
Wenhao Wang and Yi Yang.
\newblock Vidprom: A million-scale real prompt-gallery dataset for text-to-video diffusion models.
\newblock \emph{arXiv preprint arXiv:2403.06098}, 2024.

\bibitem[Wang et~al.(2019)Wang, Chan, Yu, Dong, and Change~Loy]{wang2019edvr}
Xintao Wang, Kelvin~CK Chan, Ke Yu, Chao Dong, and Chen Change~Loy.
\newblock Edvr: Video restoration with enhanced deformable convolutional networks.
\newblock In \emph{CVPRW}, pages 0--0, 2019.

\bibitem[Wang et~al.(2021)Wang, Xie, Dong, and Shan]{wang2021realesrgan}
Xintao Wang, Liangbin Xie, Chao Dong, and Ying Shan.
\newblock Real-esrgan: Training real-world blind super-resolution with pure synthetic data.
\newblock In \emph{ICCV}, pages 1905--1914, 2021.

\bibitem[Wang et~al.(2023{\natexlab{a}})Wang, Chen, Ma, Zhou, Huang, Wang, Yang, He, Yu, Yang, et~al.]{wang2023lavie}
Yaohui Wang, Xinyuan Chen, Xin Ma, Shangchen Zhou, Ziqi Huang, Yi Wang, Ceyuan Yang, Yinan He, Jiashuo Yu, Peiqing Yang, et~al.
\newblock Lavie: High-quality video generation with cascaded latent diffusion models.
\newblock \emph{arXiv preprint arXiv:2309.15103}, 2023{\natexlab{a}}.

\bibitem[Wang et~al.(2023{\natexlab{b}})Wang, He, Li, Li, Yu, Ma, Li, Chen, Chen, Wang, et~al.]{wang2023internvid}
Yi Wang, Yinan He, Yizhuo Li, Kunchang Li, Jiashuo Yu, Xin Ma, Xinhao Li, Guo Chen, Xinyuan Chen, Yaohui Wang, et~al.
\newblock Internvid: A large-scale video-text dataset for multimodal understanding and generation.
\newblock \emph{arXiv preprint arXiv:2307.06942}, 2023{\natexlab{b}}.

\bibitem[Wang et~al.(2004)Wang, Bovik, Sheikh, and Simoncelli]{wang2004image}
Zhou Wang, Alan~C Bovik, Hamid~R Sheikh, and Eero~P Simoncelli.
\newblock Image quality assessment: from error visibility to structural similarity.
\newblock \emph{IEEE TIP}, 13\penalty0 (4):\penalty0 600--612, 2004.

\bibitem[Woo et~al.(2018)Woo, Park, Lee, and Kweon]{woo2018cbam}
Sanghyun Woo, Jongchan Park, Joon-Young Lee, and In~So Kweon.
\newblock Cbam: Convolutional block attention module.
\newblock In \emph{ECCV}, pages 3--19, 2018.

\bibitem[Wu et~al.(2023)Wu, Zhang, Liao, Chen, Hou, Wang, Sun, Yan, and Lin]{wu2023dover}
Haoning Wu, Erli Zhang, Liang Liao, Chaofeng Chen, Jingwen~Hou Hou, Annan Wang, Wenxiu~Sun Sun, Qiong Yan, and Weisi Lin.
\newblock Exploring video quality assessment on user generated contents from aesthetic and technical perspectives.
\newblock In \emph{ICCV}, 2023.

\bibitem[Wu et~al.(2024)Wu, Yang, Sun, Zhang, Li, and Zhang]{wu2024seesr}
Rongyuan Wu, Tao Yang, Lingchen Sun, Zhengqiang Zhang, Shuai Li, and Lei Zhang.
\newblock Seesr: Towards semantics-aware real-world image super-resolution.
\newblock In \emph{CVPR}, pages 25456--25467, 2024.

\bibitem[Wu et~al.(2022)Wu, Wang, Li, and Shan]{wu2022animesr}
Yanze Wu, Xintao Wang, Gen Li, and Ying Shan.
\newblock Animesr: Learning real-world super-resolution models for animation videos.
\newblock \emph{NeurIPS}, 35:\penalty0 11241--11252, 2022.

\bibitem[Xu et~al.(2021)Xu, Xu, Li, Wang, Sun, and Cheng]{xu2021temporal}
Gang Xu, Jun Xu, Zhen Li, Liang Wang, Xing Sun, and Ming-Ming Cheng.
\newblock Temporal modulation network for controllable space-time video super-resolution.
\newblock In \emph{CVPR}, pages 6388--6397, 2021.

\bibitem[Xue et~al.(2019)Xue, Chen, Wu, Wei, and Freeman]{xue2019video}
Tianfan Xue, Baian Chen, Jiajun Wu, Donglai Wei, and William~T Freeman.
\newblock Video enhancement with task-oriented flow.
\newblock \emph{IJCV}, 127:\penalty0 1106--1125, 2019.

\bibitem[Yang et~al.(2023)Yang, Wu, Ren, Xie, and Zhang]{yang2023pixel}
Tao Yang, Rongyuan Wu, Peiran Ren, Xuansong Xie, and Lei Zhang.
\newblock Pixel-aware stable diffusion for realistic image super-resolution and personalized stylization.
\newblock \emph{arXiv preprint arXiv:2308.14469}, 2023.

\bibitem[Yang et~al.(2021)Yang, Xiang, Zeng, and Zhang]{yang2021realvsr}
Xi Yang, Wangmeng Xiang, Hui Zeng, and Lei Zhang.
\newblock Real-world video super-resolution: A benchmark dataset and a decomposition based learning scheme.
\newblock In \emph{ICCV}, pages 4781--4790, 2021.

\bibitem[Yang et~al.(2024{\natexlab{a}})Yang, He, Ma, and Zhang]{yang2023mgldvsr}
Xi Yang, Chenhang He, Jianqi Ma, and Lei Zhang.
\newblock Motion-guided latent diffusion for temporally consistent real-world video super-resolution.
\newblock 2024{\natexlab{a}}.

\bibitem[Yang et~al.(2024{\natexlab{b}})Yang, Teng, Zheng, Ding, Huang, Xu, Yang, Hong, Zhang, Feng, et~al.]{yang2024cogvideox}
Zhuoyi Yang, Jiayan Teng, Wendi Zheng, Ming Ding, Shiyu Huang, Jiazheng Xu, Yuanming Yang, Wenyi Hong, Xiaohan Zhang, Guanyu Feng, et~al.
\newblock Cogvideox: Text-to-video diffusion models with an expert transformer.
\newblock \emph{arXiv preprint arXiv:2408.06072}, 2024{\natexlab{b}}.

\bibitem[Yi et~al.(2019)Yi, Wang, Jiang, Jiang, and Ma]{yi2019progressive}
Peng Yi, Zhongyuan Wang, Kui Jiang, Junjun Jiang, and Jiayi Ma.
\newblock Progressive fusion video super-resolution network via exploiting non-local spatio-temporal correlations.
\newblock In \emph{ICCV}, pages 3106--3115, 2019.

\bibitem[Yu et~al.(2024)Yu, Gu, Li, Hu, Kong, Wang, He, Qiao, and Dong]{yu2024scaling}
Fanghua Yu, Jinjin Gu, Zheyuan Li, Jinfan Hu, Xiangtao Kong, Xintao Wang, Jingwen He, Yu Qiao, and Chao Dong.
\newblock Scaling up to excellence: Practicing model scaling for photo-realistic image restoration in the wild.
\newblock In \emph{CVPR}, pages 25669--25680, 2024.

\bibitem[Yuan et~al.(2024)Yuan, Baek, Xu, Tov, and Fei]{yuan2024inflation}
Xin Yuan, Jinoo Baek, Keyang Xu, Omer Tov, and Hongliang Fei.
\newblock Inflation with diffusion: Efficient temporal adaptation for text-to-video super-resolution.
\newblock In \emph{Proceedings of the IEEE/CVF Winter Conference on Applications of Computer Vision}, pages 489--496, 2024.

\bibitem[Yue et~al.(2024)Yue, Wang, and Loy]{yue2024resshift}
Zongsheng Yue, Jianyi Wang, and Chen~Change Loy.
\newblock Resshift: Efficient diffusion model for image super-resolution by residual shifting.
\newblock \emph{NeurIPS}, 36, 2024.

\bibitem[Zhang et~al.(2015)Zhang, Zhang, and Bovik]{zhang2015ilniqe}
Lin Zhang, Lei Zhang, and Alan~C Bovik.
\newblock A feature-enriched completely blind image quality evaluator.
\newblock \emph{IEEE TIP}, 24\penalty0 (8):\penalty0 2579--2591, 2015.

\bibitem[Zhang et~al.(2023{\natexlab{a}})Zhang, Rao, and Agrawala]{zhang2023adding}
Lvmin Zhang, Anyi Rao, and Maneesh Agrawala.
\newblock Adding conditional control to text-to-image diffusion models.
\newblock In \emph{ICCV}, pages 3836--3847, 2023{\natexlab{a}}.

\bibitem[Zhang et~al.(2018)Zhang, Isola, Efros, Shechtman, and Wang]{zhang2018lpips}
Richard Zhang, Phillip Isola, Alexei~A Efros, Eli Shechtman, and Oliver Wang.
\newblock The unreasonable effectiveness of deep features as a perceptual metric.
\newblock In \emph{CVPR}, pages 586--595, 2018.

\bibitem[Zhang et~al.(2023{\natexlab{b}})Zhang, Wang, Zhang, Zhao, Yuan, Qin, Wang, Zhao, and Zhou]{zhang2023i2vgen}
Shiwei Zhang, Jiayu Wang, Yingya Zhang, Kang Zhao, Hangjie Yuan, Zhiwu Qin, Xiang Wang, Deli Zhao, and Jingren Zhou.
\newblock I2vgen-xl: High-quality image-to-video synthesis via cascaded diffusion models.
\newblock \emph{arXiv preprint arXiv:2311.04145}, 2023{\natexlab{b}}.

\bibitem[Zhang and Yao(2024)]{zhang2024realviformer}
Yuehan Zhang and Angela Yao.
\newblock Realviformer: Investigating attention for real-world video super-resolution.
\newblock \emph{ECCV}, 2024.

\bibitem[Zhao et~al.(2024)Zhao, Cai, Dong, and Hu]{zhao2024wavelet}
Chen Zhao, Weiling Cai, Chenyu Dong, and Chengwei Hu.
\newblock Wavelet-based fourier information interaction with frequency diffusion adjustment for underwater image restoration.
\newblock In \emph{Proceedings of the IEEE/CVF Conference on Computer Vision and Pattern Recognition}, pages 8281--8291, 2024.

\bibitem[Zhou et~al.(2024)Zhou, Yang, Wang, Luo, and Loy]{zhou2024upscale}
Shangchen Zhou, Peiqing Yang, Jianyi Wang, Yihang Luo, and Chen~Change Loy.
\newblock Upscale-a-video: Temporal-consistent diffusion model for real-world video super-resolution.
\newblock In \emph{CVPR}, pages 2535--2545, 2024.

\end{thebibliography}
}

\newpage
\appendix

\section{Perception-Distortion Trade-Off}

\begin{figure}[b]
    \centering
    \includegraphics[width=1\linewidth]{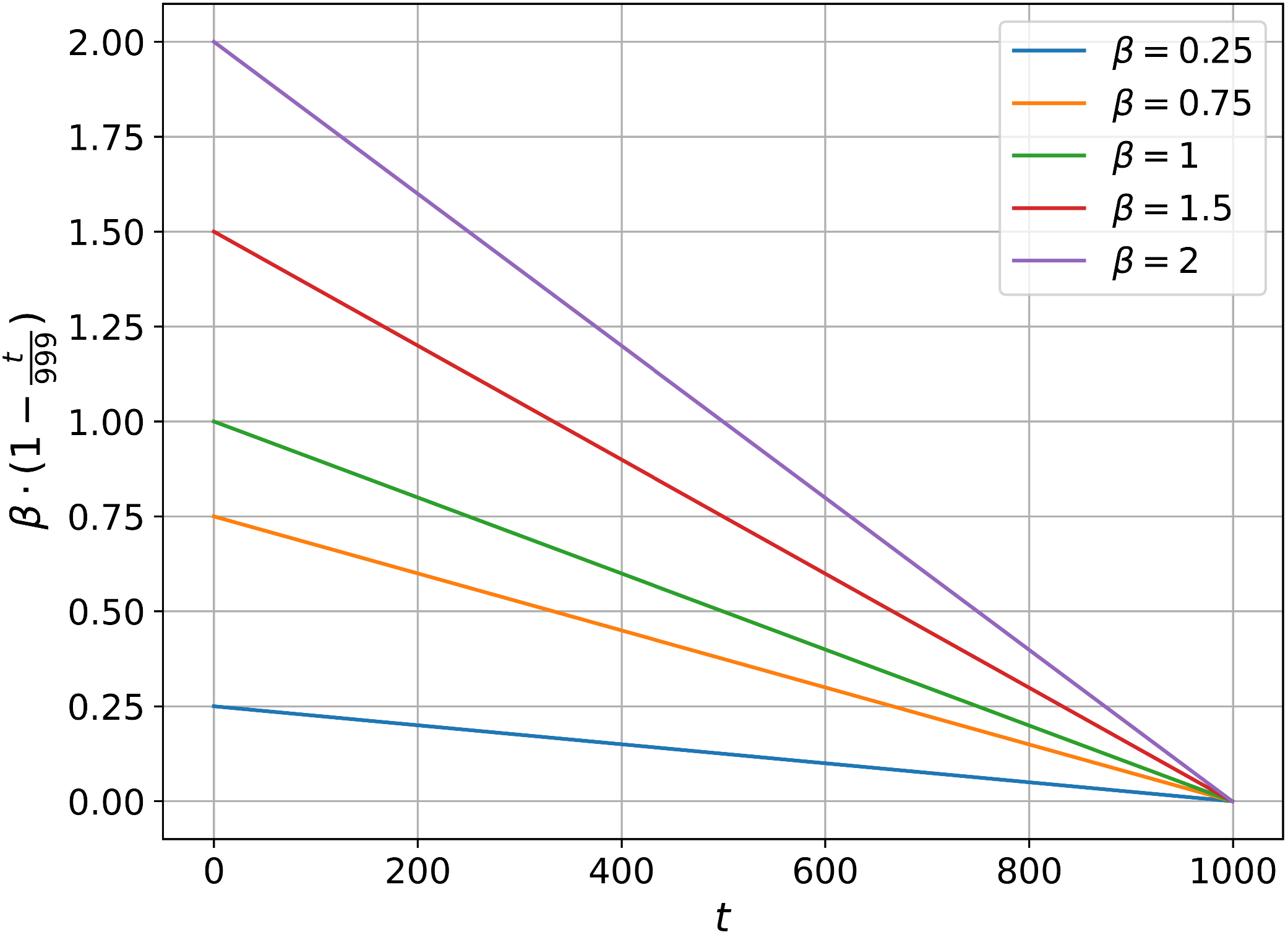}
    \caption{Ablation on $b(t)$. Higher hyper-parameter $\beta$ produces results with greater fidelity, while lower $\beta$ emphasizes more perceptual quality.}
    \label{fig:bt_ablation}
\end{figure}

The trade-off between perception and distortion \cite{blau2018perception} is a widely recognized challenge in the super-resolution domain. Thanks to our \textit{DF Loss}, our method can easily control the model to favor either fidelity or perceptual quality in the generated results. We can adjust the hyper-parameter $\beta$ in the $b(t)$ to achieve this goal. The total loss in our~\name~is:
\begin{equation}
    \mathcal{L}_{total} = \mathcal{L}_{v} + b(t)\mathcal{L}_{DF},
\end{equation}
The $b(t)$ can be written as follows:
\begin{equation}
    b(t) = \beta \cdot (1 - \frac{t}{t_{max}}),
\end{equation}
Where \( t \) is the timestep and \( \beta \) is the hyper-parameter that adjusts the weight between \( \mathcal{L}_v \) and \( \mathcal{L}_{DF} \), which we set to 1 by default. From equations (1) and (2), we can observe that a larger \( \beta \) increases the weight of the DF loss at each timestep, thereby further enhancing the fidelity of the results. In contrast, a smaller \( \beta \) reduces the influence of the DF loss at each timestep, allowing the v-prediction loss to have a greater impact and produce more perceptual results. The $b(t)$ - $t$ curves under different $\beta$ are shown in Figure \ref{fig:bt_ablation}. 

We conduct experiments under these settings to demonstrate the ability to achieve the perception-distortion trade-off. The quantitative results are shown in Table \ref{tab:beta_ablation}. From Table \ref{tab:beta_ablation}, we can observe that increasing $\beta$ improves the PSNR and $E_{warp}^*$, leading to better fidelity. Conversely, decreasing $\beta$ reduces the LPIPS score, indicating better perceptual quality.

\begin{table}[]
    \centering
    \caption{Qualitative comparison under different $\beta$ of $b(t)$.}
    \begin{tabular}{c|ccc}
    \hline
    $\beta$ & PSNR$\uparrow$ & LPIPS$\downarrow$ & $E_{warp}^*\downarrow$ \\ \hline
       0.25 & 23.55 & \textbf{0.1825} & 2.88  \\
       0.75 & 23.76 & 0.1842 & 2.74  \\
       1.0 & 23.91 & 0.1885 & 2.68  \\
       1.5 & 24.08 & 0.2272 & 2.53  \\
       2.0 & \textbf{24.41} & 0.3339 & \textbf{2.21} \\ \hline
    \end{tabular}
    \label{tab:beta_ablation}
\end{table}

\section{More Results}
\subsection{User Study}
\begin{figure*}[]
    \centering
    \includegraphics[width=\linewidth]{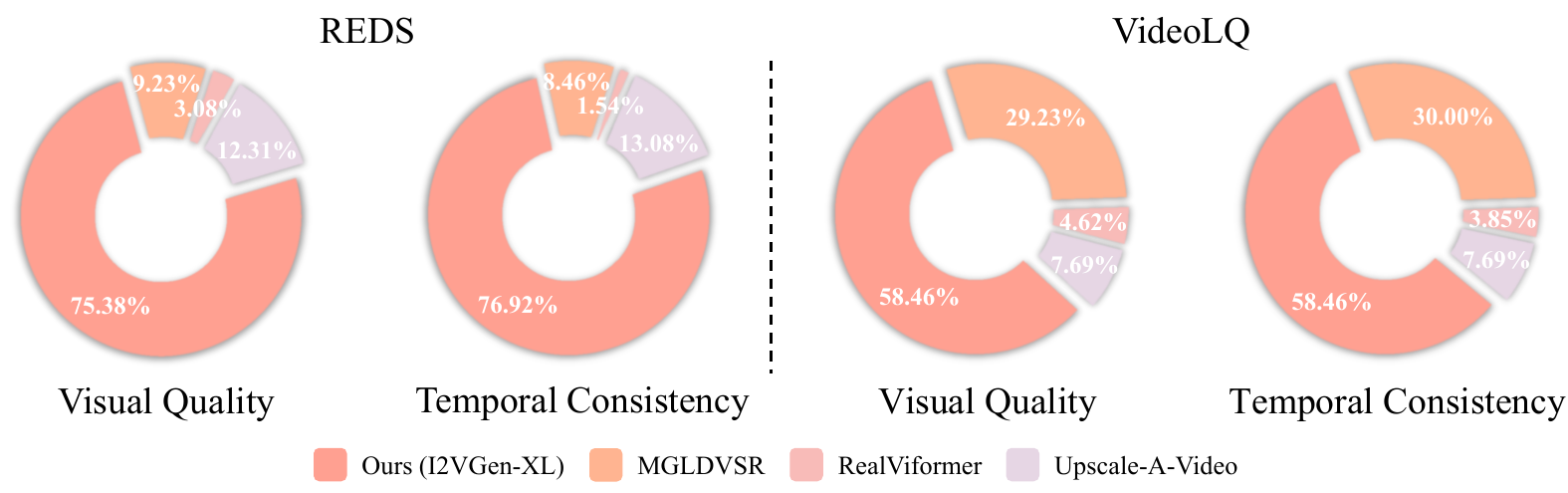}
    \caption{User study results. Our \name\ is preferred by human evaluators for both visual quality and temporal consistency.}
    \label{fig:user_study}
\end{figure*}

To find the human-preferred results between our~\name~and other state-of-the-art methods, we conduct a user study that evaluate the results on both real-world and synthetic datasets. Specifically, we use the real-world dataset VideoLQ \cite{chan2022investigating} and the synthetic dataset REDS30 \cite{nah2019ntire}. We select two image-diffusion-model-based methods, Upscale-A-Video \cite{zhou2024upscale} and MGLD-VSR \cite{yang2023mgldvsr}; and one GAN-based method, RealViformer \cite{zhang2024realviformer} for comparison. 
We invite 12 evaluators to participate in the user study. For each evaluator, we randomly select 10 videos from each dataset and present four results: one from our~\name~and three from the compared methods. The evaluators were asked to choose which result had the best visual quality and temporal consistency.
The results of the user study are depicted in Figure \ref{fig:user_study}, indicating that our \name\ is preferred by most human evaluators for both visual quality and temporal consistency.

\begin{figure*}
    \centering
    \includegraphics[width=1\linewidth]{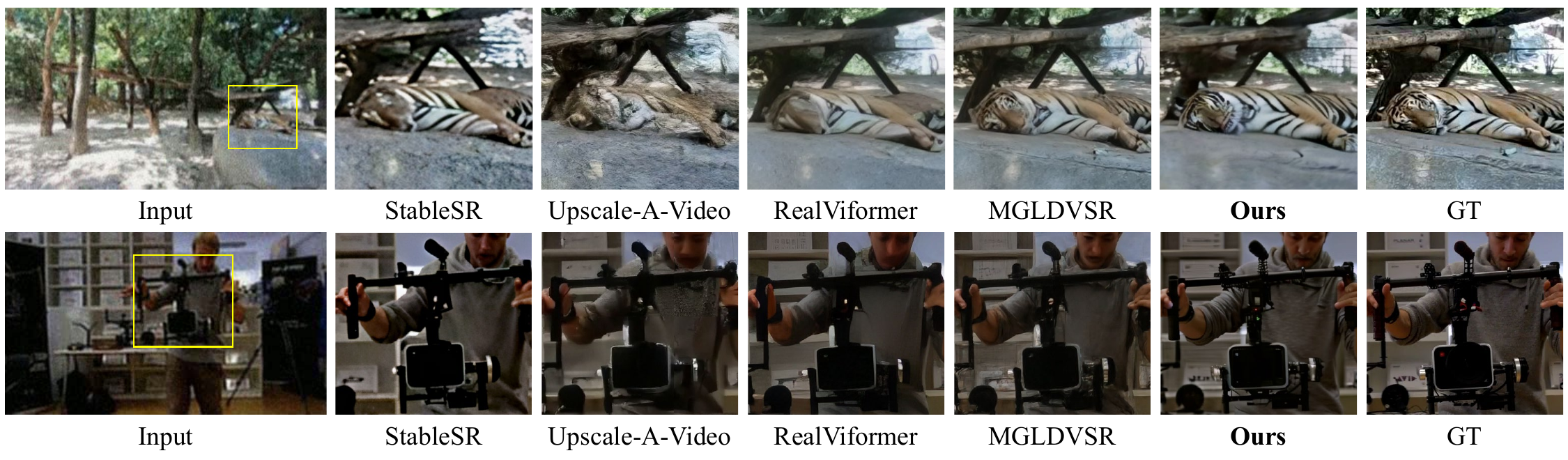}
    \caption{Qualitative comparisons on synthetic datasets. Our~\name~generates more detailed and realistic results. \textbf{(Zoom-in for best view)}}
    \label{fig:synthetic_comparison}
\end{figure*}

\begin{figure*}
    \centering
    \includegraphics[width=1\linewidth]{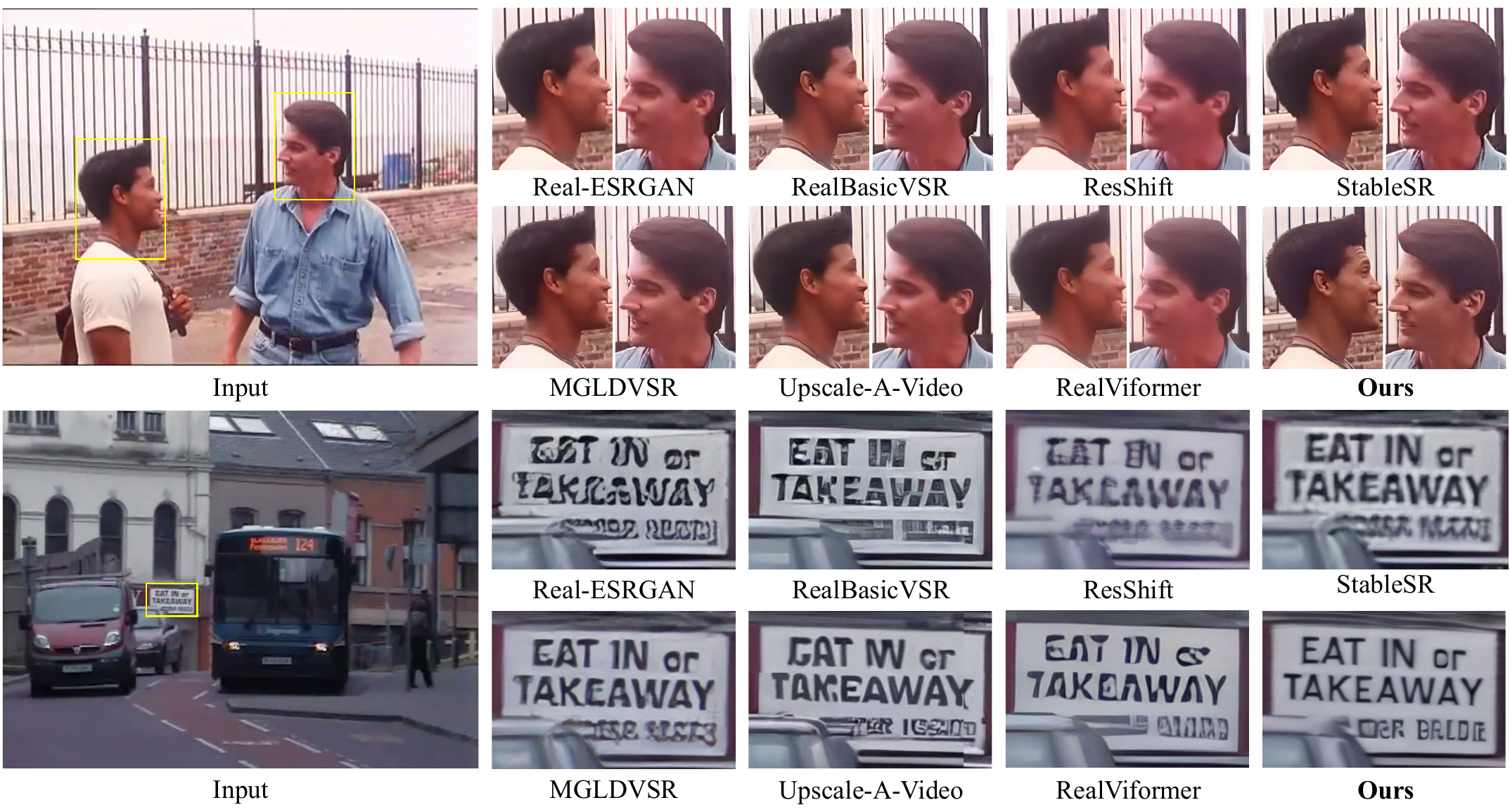}
    \caption{Qualitative comparisons on real-world datasets. Our~\name~produces the clearest facial details and the most accurate text structure. \textbf{(Zoom-in for best view)}}
    \label{fig:realworld_comparison}
\end{figure*}

\begin{figure*}
    \centering
    \includegraphics[width=1\linewidth]{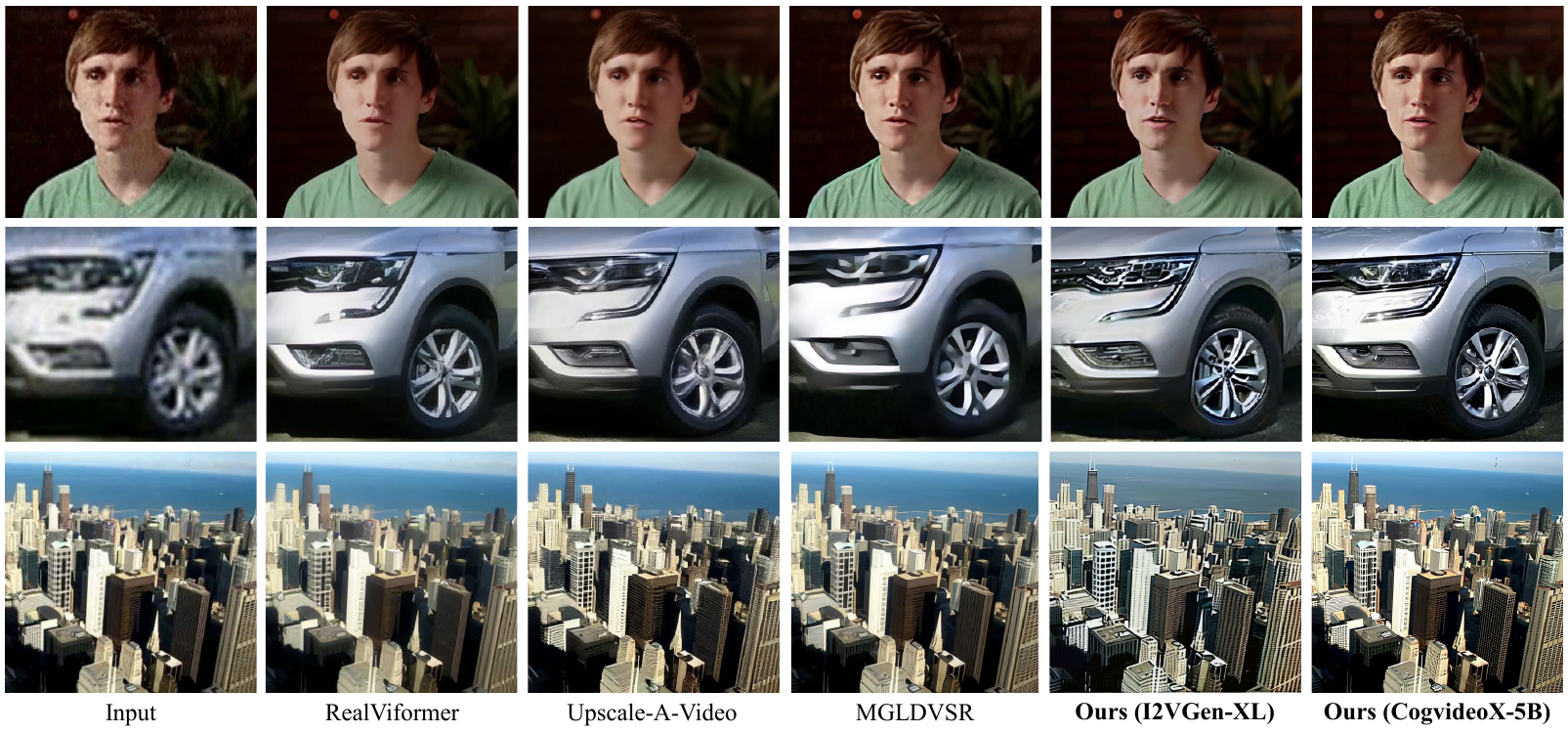}
    \caption{Qualitative comparisons on synthetic and real-world datasets with larger T2V models. Scaling up the T2V model enhances detail and realism in video super-resolution results. \textbf{(Zoom-in for best view)}}
    \label{fig:scale_up}
\end{figure*}

\subsection{Qualitative Comparisons}
We provide more visual comparisons on synthetic and real-world datasets in Figure \ref{fig:synthetic_comparison} and Figure \ref{fig:realworld_comparison} to further highlight our advantages in spatial quality. These results clearly demonstrate that our method preserves richer details and achieves greater realism.
To demonstrate the impact of scaling up with larger text-to-video (T2V) models, we present additional results in Figure \ref{fig:scale_up}. It is evident that scaling up the T2V model further improves the restoration effect, indicating that a large and robust T2V model can serve as a strong base model for video super-resolution.

\subsection{Video Demo}
We provide a demo video \href{https://youtu.be/hx0zrql-SrU}{\textcolor{red}{[STAR-demo.mp4]}} in the supplementary material, showcasing the temporal and spatial advantages of our proposed~\name~more intuitively. This video includes additional results and comparisons on synthetic, real-world, and AIGC videos.

\end{document}